\PassOptionsToPackage{table}{xcolor}

\documentclass{article}
\usepackage[preprint]{colm2026_conference}

\usepackage[utf8]{inputenc}
\usepackage[T1]{fontenc}
\usepackage{microtype}
\usepackage[bookmarksopen,
bookmarksdepth=2,
breaklinks=true,
colorlinks=true,
linkcolor=linkcolour,
citecolor=citecolour,
filecolor=urlcolour,
urlcolor=urlcolour]{hyperref}
\usepackage{url}
\usepackage{booktabs}
\usepackage{graphicx}
\usepackage{wrapfig}
\usepackage{subcaption}
\usepackage{amsmath,amsfonts,amssymb}
\usepackage{float}
\usepackage{lineno}
\usepackage{titletoc}
\usepackage{xspace}
\usepackage{multirow}

\usepackage{arydshln}

\usepackage{listings}
\newcommand{\github}{\raisebox{-1.5pt}{\includegraphics[height=1.05em]{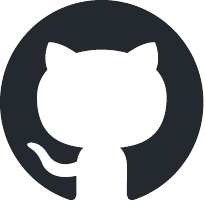}}\xspace}
\newcommand{\notion}{\raisebox{-1.5pt}{\includegraphics[height=1.05em]{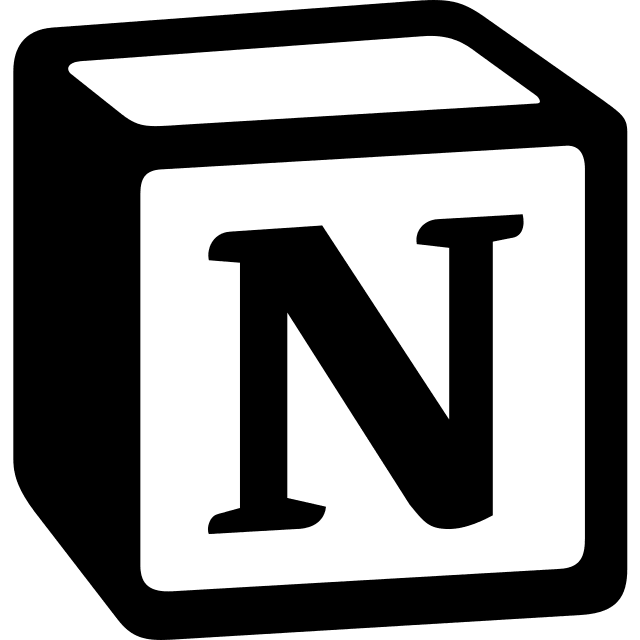}}\xspace}

\definecolor{darkblue}{rgb}{0, 0, 0.5}
\hypersetup{colorlinks=true, citecolor=darkblue, linkcolor=darkblue, urlcolor=darkblue}
\graphicspath{{figures/}}

\title{Revisiting On-Policy Distillation:\\Empirical Failure Modes and Simple Fixes}

\author{%
\normalfont
\begin{tabular}[t]{@{}l@{}}
\textbf{Yuqian Fu}\textsuperscript{1,2}\thanks{Equal contribution. $^\dag$Corresponding authors. $^\ddag$Work in progress.}\quad
\textbf{Haohuan Huang}\textsuperscript{1,2}\footnotemark[1]\quad
\textbf{Kaiwen Jiang}\textsuperscript{1,2}\quad
\textbf{Jiacai Liu}\textsuperscript{3}\\[0.25em]
\textbf{Zhuo Jiang}\textsuperscript{4}\quad
\textbf{Yuanheng Zhu}\textsuperscript{1,2}\footnotemark[2]\quad
\textbf{Dongbin Zhao}\textsuperscript{1,2}\\[0.3em]
\small
\textsuperscript{1}\,State Key Laboratory of Multimodal Artificial Intelligence Systems, CASIA\\[0.15em]
\textsuperscript{2}\,School of Artificial Intelligence, UCAS\enspace
\textsuperscript{3}\,Fudan University\enspace
\textsuperscript{4}\,Independent Researcher\\[0.1em]
\texttt{\{fuyuqian2022, yuanheng.zhu\}@ia.ac.cn}
\end{tabular}%
}
\begin{document}

\ifcolmsubmission
\linenumbers
\fi

\maketitle
\vspace{-1em}
\begin{abstract}
On-policy distillation (OPD) is increasingly used in LLM post-training because it can leverage a teacher model to provide dense supervision on student rollouts. The standard implementation, however, usually reduces distribution matching to a sampled-token log-ratio, which can make the learning signal fragile on long rollouts whose prefixes drift away from the teacher's typical support. We revisit this formulation from both theoretical and implementation perspectives. Theoretically, token-level OPD is biased relative to sequence-level reverse-KL minimization, but admits a substantially tighter worst-case variance bound; a controlled synthetic study further shows that stronger future-reward coupling increases gradient variance and destabilizes training. Empirically, we identify three failure modes of sampled-token OPD: imbalanced token-level supervision, unreliable teacher guidance on student-generated prefixes, and tokenizer or special-token mismatch. These findings motivate teacher top-$K$ local support matching, a truncated reverse-KL objective that compares teacher and student distributions over a teacher-supported token set at each prefix, together with top-$p$ rollout sampling and special-token masking. Across single-task reasoning and multi-task benchmarks spanning agentic and reasoning settings, this objective improves optimization stability and yields a \textbf{+19.8\%} performance gain over standard sampled-token OPD baselines, providing a practical recipe for more stable on-policy distillation.

\begin{center}
\renewcommand{\arraystretch}{1.2}
\begin{tabular}{rcrc}
    \github &  \href{https://github.com/hhh675597/revisiting_opd}{\textbf{Code}} &
    \notion & \href{https://yuqianfu.notion.site/revisiting-opd}{\textbf{Blog}}
\end{tabular}
\end{center}

\end{abstract}

\section{Introduction}
\label{sec:introduction}

On-policy distillation (OPD) is becoming an increasingly common component of LLM post-training, especially for reasoning and agentic models. Recent public reports from Thinking Machines Lab~\citep{tmlopd2025}, Qwen3~\citep{qwen3_2025}, MiMo-V2-Flash~\citep{mimo_v2_flash_2026}, and GLM-5~\citep{glm5_2026} suggest a broader shift toward supervision on model-generated trajectories, or closely related on-policy variants, alongside off-policy distillation and reinforcement learning. By training on student rollouts while evaluating them with a stronger teacher, OPD combines on-policy data collection with dense token-level feedback at relatively low cost~\citep{opd2023,minillm2023}. This profile is attractive in practical post-training pipelines, where training efficiency matters and models often need to combine or recover capabilities across domains and training stages~\citep{mimo_v2_flash_2026,glm5_2026,wang2026mix,deepseekai2026deepseekv4}.

Most OPD implementations in current LLM pipelines use a token-level estimator, even though it is biased relative to sequence-level reverse-KL~\citep{tmlopd2025}. A basic reason is that sequence-level objectives couple each token update to many future rewards, which can make optimization substantially noisier in long-horizon settings. We make this trade-off explicit: token-level OPD removes future-reward coupling and is therefore biased, yet it admits more favorable worst-case variance scaling. Our toy experiment shows the same pattern empirically, with stronger future coupling leading to higher gradient variance and less stable optimization. This suggests a practical design principle for long-horizon training: \textit{keep supervision token-level to control variance.}

In current LLM pipelines, this token-level objective is usually instantiated by sampled-token comparison: at each training step, the update is driven by the teacher--student log-probability difference on the sampled token~\citep{tmlopd2025,mimo_v2_flash_2026}. This implementation is simple and efficient, but its learning signal can become brittle once rollouts grow long. \citet{minillm2023} report degraded outputs such as repetition, which is consistent with our observations. Recent work also reports entropy collapse under sampled-token OPD in certain cases~\citep{ko2026reopold,jin2026eopd}. More recently, full-vocabulary distillation has been reported to outperform sampled-token variants in some settings~\citep{zhao2026opsd,deepseekai2026deepseekv4}, suggesting that the one-token formulation can leave useful teacher information unused.

Our empirical analysis shows that this brittleness stems from several recurring failure modes. In particular, we identify two \textbf{objective-level} issues: the one-token signal is often highly imbalanced, and teacher guidance can become unreliable on student-generated prefixes. We also observe an additional \textbf{implementation-level} issue from tokenizer or special-token mismatch, which can further distort one-token comparisons. Taken together, these results point to a practical question: \textit{how can we retain the variance advantage of token-level OPD while making its supervision signal less brittle in practice?}

The analysis above suggests a targeted modification of the standard sampled-token objective. We replace one-token supervision with teacher top-$K$ local support matching, where teacher and student are compared over a teacher-supported token subset at each prefix rather than only on the sampled token. We implement this objective as truncated reverse-KL, together with top-$p$ rollout sampling and special-token masking. The resulting update remains local and inexpensive, while providing a less brittle training signal.

Overall, \textbf{our main contributions} are as follows:
\begin{itemize}
    \item We clarify the theoretical trade-off in OPD: token-level OPD is biased relative to sequence-level OPD, but has substantially better worst-case variance scaling with sequence length, making it attractive for long-horizon LLM post-training.
    \item We present an empirical analysis of why sampled-token OPD can be unstable in practice, highlighting two recurring objective-level issues---imbalanced one-token supervision and unreliable teacher guidance on student-generated prefixes---along with an additional implementation issue from tokenizer or special-token mismatch.
    \item We propose teacher top-$K$ local support matching as an analysis-driven revision of sampled-token OPD, implemented with truncated reverse-KL, top-$p$ rollouts, and special-token masking, and show that it yields more stable optimization and stronger empirical performance than sampled-token OPD in single-task math reasoning and multi-task agentic-plus-reasoning training.
\end{itemize}

\section{Understanding Sampled-Token OPD: Trade-offs and Failure Modes}
\label{sec:revisiting_opd}

\subsection{From reverse-KL to token-level OPD}
\label{subsec:bias_variance}

\begin{figure}[t]
    \centering
    \begin{subfigure}{\linewidth}
        \centering
        \includegraphics[width=0.75\linewidth]{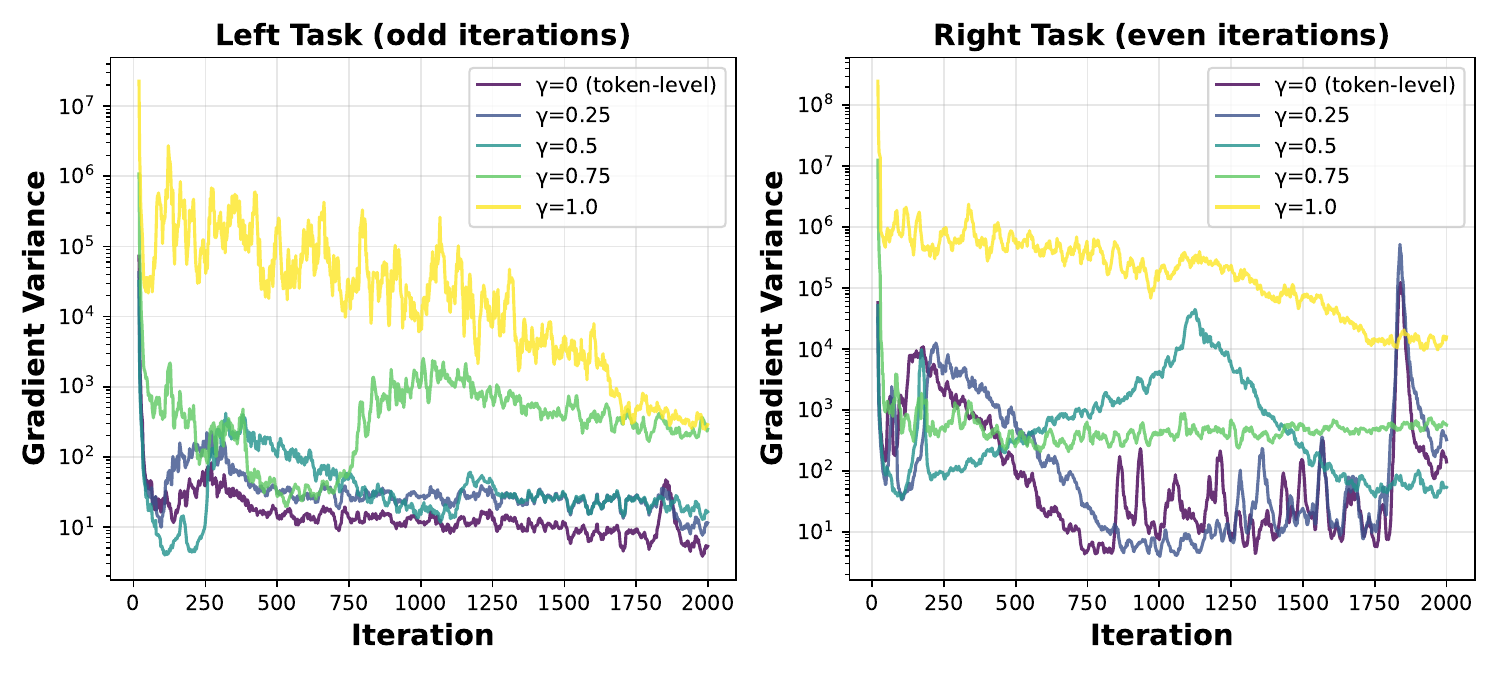}
        \caption{Gradient variance in the toy experiment. Larger $\gamma$ generally yields higher variance in both tasks.}
    \end{subfigure}
    \begin{subfigure}{\linewidth}
        \centering
        \includegraphics[width=0.75\linewidth]{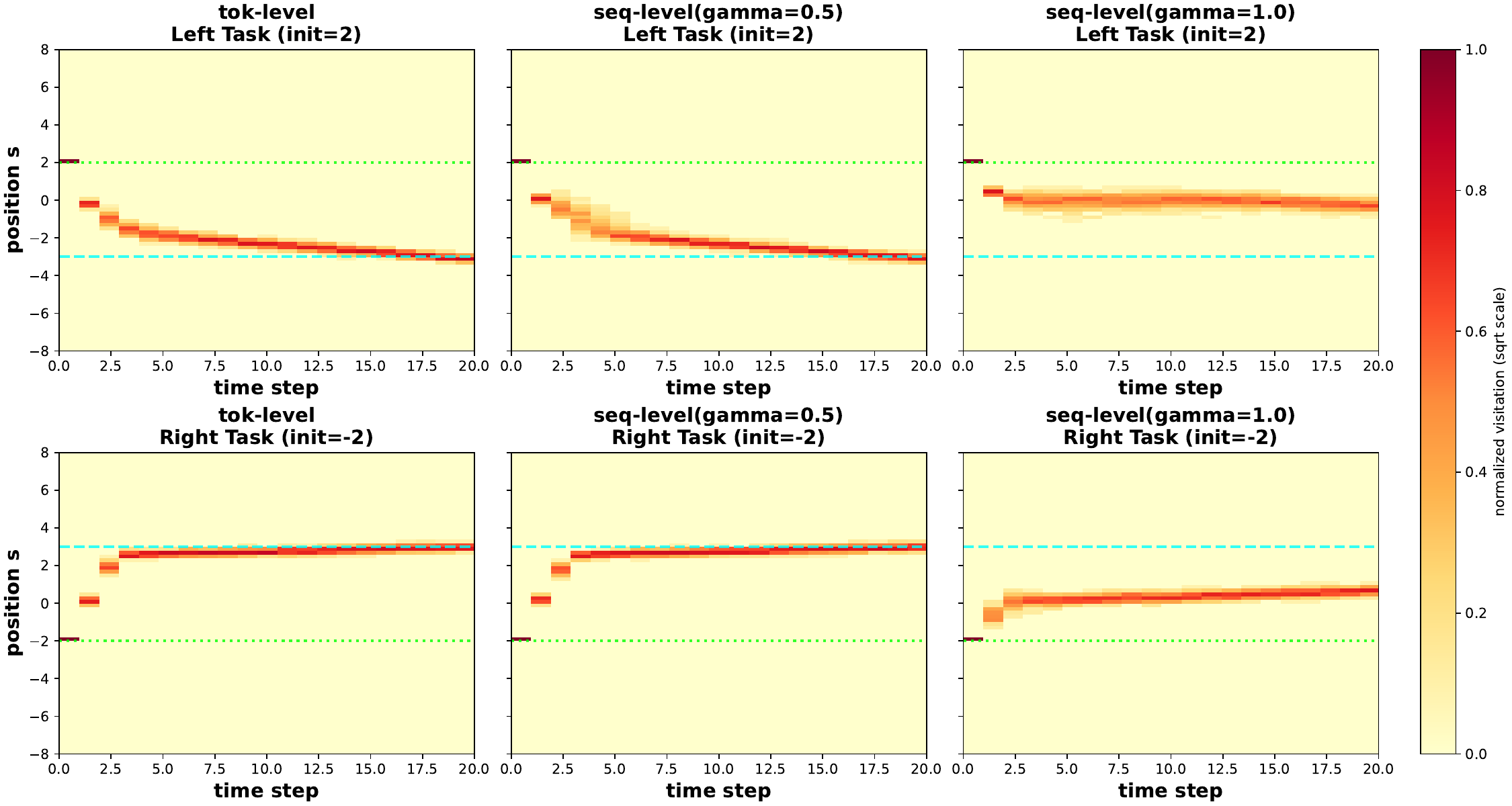}
        \caption{State visitation under $\gamma \in \{0.0, 0.5, 1.0\}$ in the toy environment. For $\gamma=1.0$, the policy model fails to consistently move toward the target, and instead exhibits drifting behavior.}
    \end{subfigure}
    \caption{Effect of increasing $\gamma$ in the toy experiment. Larger $\gamma$ yields a higher and more persistent variance regime and, in the sequence-level limit, drifting policies in state space.}
    \label{fig:toy_variance}
\end{figure}

On-policy distillation aims to transfer the capabilities of a stronger teacher model into a student model by minimizing the reverse-KL from the student to the teacher. For a prompt $x$, the OPD objective is
\[
J_{\mathrm{OPD}}(\theta)
=
\mathbb{E}_{x \sim D}\left[
D_{\mathrm{KL}}\left(\pi_\theta(\cdot \mid x)\,\|\,q(\cdot \mid x)\right)
\right],
\]
where $\pi_\theta$ and $q$ denote the student and teacher models, respectively. Its gradient can be written as
\[
\nabla_\theta J_{\mathrm{OPD}}(\theta)
=
\mathbb{E}_{x,\, y \sim \pi_\theta(\cdot \mid x)}\left[
\big(\log \pi_\theta(y \mid x)-\log q(y \mid x)\big)\,
\nabla_\theta \log \pi_\theta(y \mid x)
\right].
\]
For each decoding step $t$, let $c_t = (x, y_{<t})$ denote the prefix context, and define
\[
s_t = \nabla_\theta \log \pi_\theta(y_t \mid c_t),
\qquad
r_t = \log \frac{\pi_\theta(y_t \mid c_t)}{q(y_t \mid c_t)}.
\]
Using the autoregressive factorization
\[
\log \pi_\theta(y \mid x) - \log q(y \mid x) = \sum_{t'=1}^{T} r_{t'},
\qquad
\nabla_\theta \log \pi_\theta(y \mid x) = \sum_{t=1}^{T} s_t,
\]
we obtain the sequence-level estimator
\begin{equation}
\hat g_{\mathrm{seq}} = \sum_{t=1}^{T} \left(\sum_{t'=1}^{T} r_{t'}\right) s_t.
\label{eq:seq_opd}
\end{equation}
For $t' < t$, we have $\mathbb{E}[r_{t'} s_t] = 0$, because $r_{t'}$ depends only on the prefix before step $t$, while
\[
\mathbb{E}[s_t \mid x, y_{<t}]
=
\sum_{y_t} \pi_\theta(y_t \mid c_t)\, \nabla_\theta \log \pi_\theta(y_t \mid c_t)
= 0.
\]
The same gradient can therefore be expressed in causal reward-to-go form
\[
\mathbb{E}[\hat g_{\mathrm{seq}}]
=
\mathbb{E}\left[
\sum_{t=1}^{T}
\left(\sum_{t'=t}^{T} r_{t'}\right) s_t
\right],
\]
where each token update is coupled to all future rewards along the trajectory.

Another approximation in LLM training retains only the immediate term at each position
\begin{equation}
\hat g_{\mathrm{tok}} = \sum_{t=1}^{T} r_t s_t.
\label{eq:token_opd}
\end{equation}
We refer to Eq.~\eqref{eq:token_opd} as the token-level estimator.
This approximation removes future-reward coupling, so the update for token $y_t$ depends only on its immediate reward. Consequently, it is biased relative to the sequence-level reverse-KL estimator, but exhibits lower variance in long-horizon settings.
Under bounded rewards and bounded gradients, the worst-case variance upper bound of token-level OPD scales as $O(T^2)$, whereas the sequence-level estimator scales as $O(T^4)$. We provide a detailed derivation in Appendix~\ref{app:bias_variance}. To interpolate between these extremes, we consider the discounted return-to-go estimator
\begin{equation}
\hat g_{\gamma}
=
\sum_{t=1}^{T}
\left(\sum_{t'=t}^{T} \gamma^{t'-t} r_{t'} \right) s_t.
\qquad
\gamma \in [0,1]
\label{eq:gamma_estimator}
\end{equation}
The case $\gamma=0$ recovers token-level OPD, while $\gamma=1$ recovers the causal sequence-level estimator. We further validate this trade-off in a two-task toy experiment (Figure~\ref{fig:toy_variance}): stronger future coupling leads to substantially higher gradient variance and less stable optimization. This motivates our focus on token-level supervision in the remainder of the paper, where the main question becomes how to improve its local training signal in practical LLM settings. Additional experimental details are provided in Appendix~\ref{app:toy}.

\subsection{Why sampled-token OPD is brittle in practice}
\label{subsec:brittle}

Although token-level OPD is appealing from a bias--variance perspective, the standard sampled-token formulation can be brittle in practice. We isolate three failure modes: (1) a highly imbalanced token-level distillation signal, (2) unreliable teacher guidance on student-generated prefixes, and (3) distortions introduced by tokenizer or special-token mismatch.
These observations come from sampled-token OPD experiments on math reasoning, using Qwen2.5-7B-Instruct~\citep{qwen2025qwen25} as the student and OpenThinker3-7B~\citep{openthoughts2025}, an SFT model built on Qwen2.5-7B-Instruct, as the teacher.

\begin{figure}[t]
\centering
\includegraphics[width=0.35\linewidth]{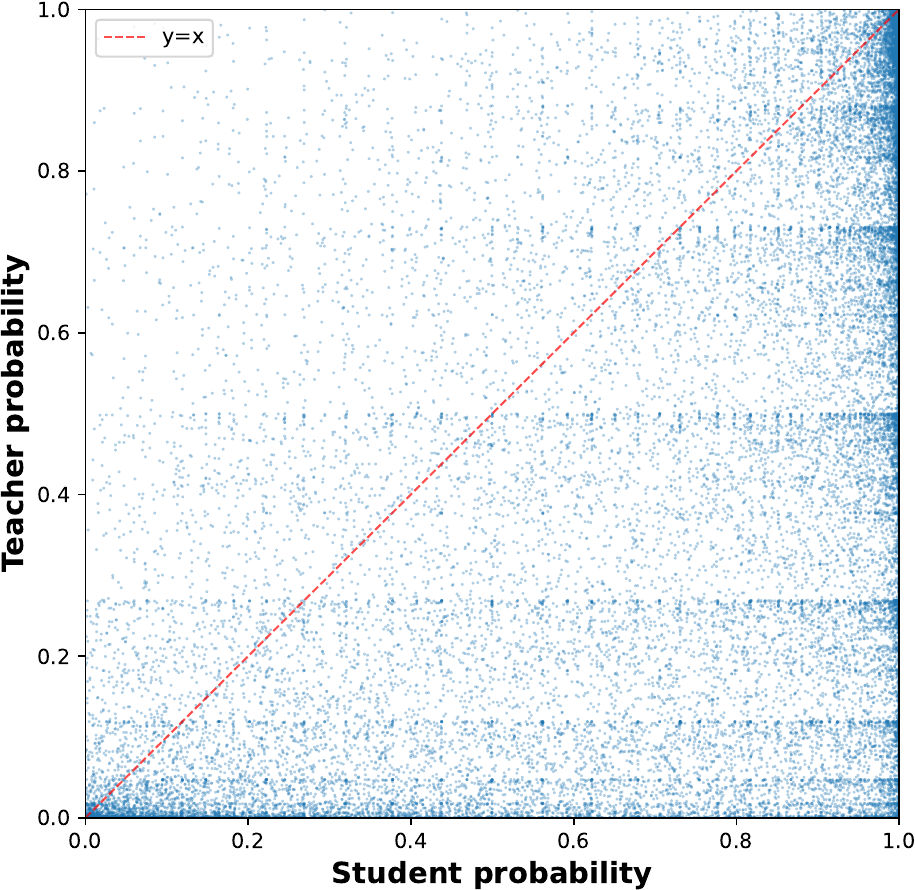}
\caption{Scatter of token probabilities (teacher vs. student) at the sampled-token OPD first iteration. The sampled-token signal is heavily skewed toward penalizing the current token.}
\label{fig:imbalance_reward}
\vspace{-1em}
\end{figure}

\paragraph{A highly imbalanced sampled-token signal.}
In sampled-token OPD, the update at step $t$ is driven by the log-ratio on a single sampled token:
\[
\log q(y_t \mid c_t) - \log \pi_\theta(y_t \mid c_t).
\]
Negative rewards arise whenever the student assigns higher probability to a sampled token than the teacher.
As shown in Figure~\ref{fig:imbalance_reward}, most sampled tokens receive negative reward.
This leaves optimization dominated by a small subset of locally positive tokens, so training becomes sensitive to high-frequency fillers and short continuations that can receive favorable local scores while contributing little to trajectory-level quality.

\begin{figure}[t]
    \centering
    \includegraphics[width=0.75\linewidth]{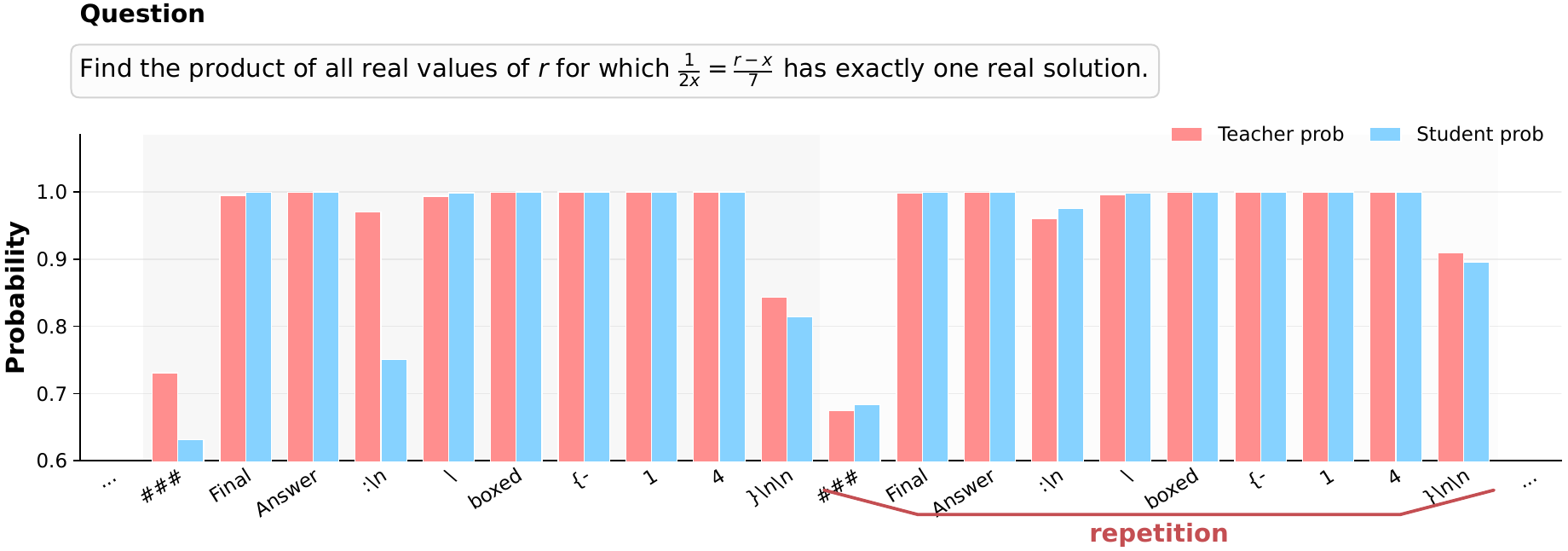}
    \caption{Example of unreliable teacher guidance. The student falls into a repetition loop, yet the teacher remains locally aligned with the student on repeated tokens, indicating that sampled-token OPD may fail to penalize this behavior.}
    \label{fig:repeat}
    \vspace{-1em}
\end{figure}

\paragraph{The teacher signal can become unreliable on student-generated prefixes.}
Sampled-token OPD assumes that teacher probability on a student-generated token is a useful proxy for trajectory quality.
This proxy becomes unreliable on prefixes that are common under the student but uncommon for the teacher. In such regions, tokens with high teacher probability can remain rewarded even after the trajectory has drifted into repetition, self-resetting reasoning, or other meaningless continuations (Figure~\ref{fig:repeat}; Appendix~\ref{app:case_study}).
This creates an objective mismatch between token-level teacher agreement and trajectory-level quality.

\begin{figure}[t]
\centering
\includegraphics[width=0.75\linewidth]{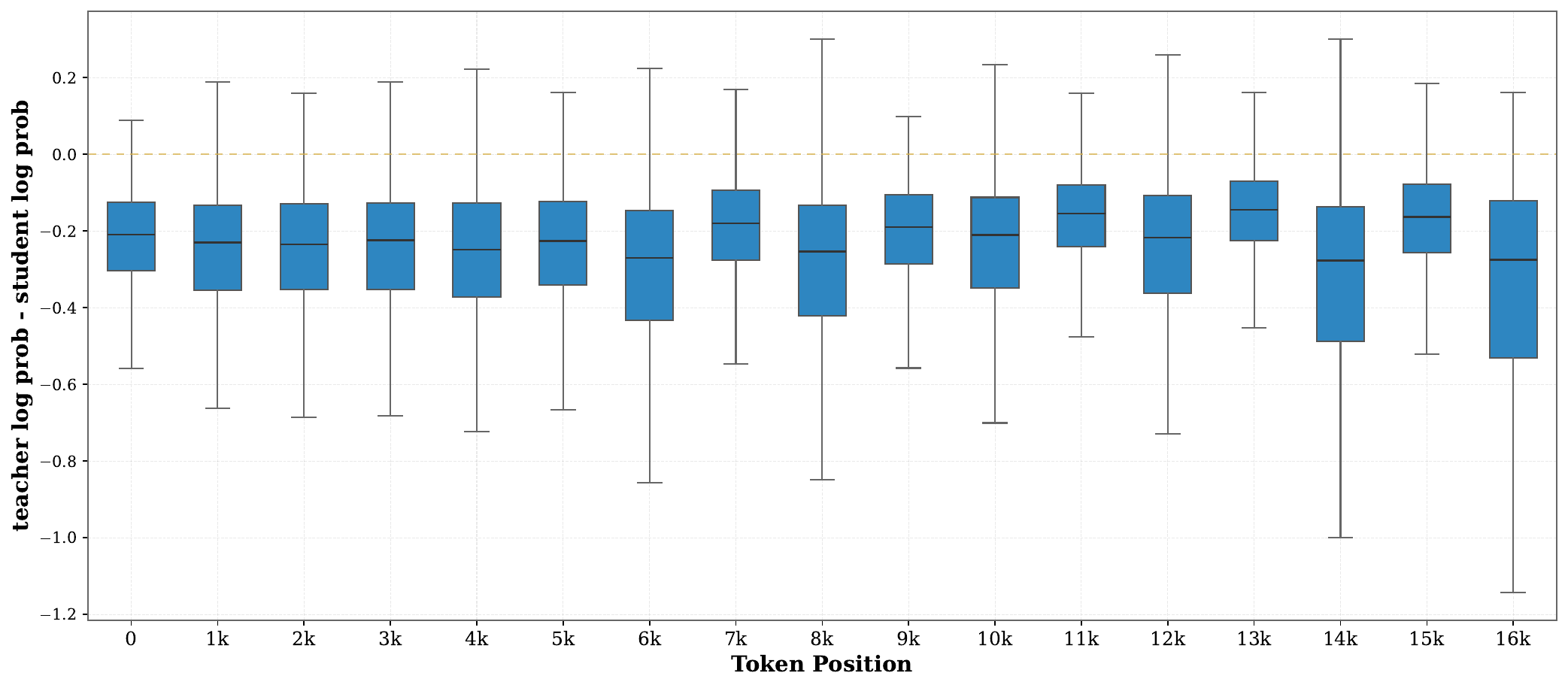}
\caption{Distribution of teacher--student log-probability gaps across token positions. Several later position buckets exhibit wider lower tails and more extreme values, indicating a noisier teacher signal on long-horizon student-generated rollouts.}
\label{fig:teacher_signal_variability}
\end{figure}

We hypothesize that two factors amplify this issue: sharp teacher distributions, where modest teacher--student mismatch can produce large log-ratio values, and growing teacher--student divergence along long rollouts. Consistent with this view, Figure~\ref{fig:teacher_signal_variability} shows that the distribution of teacher--student gaps becomes wider later in the sequence.

\paragraph{Tokenizer and special-token mismatch.}
Sampled-token OPD compares the exact token generated by the student using the teacher distribution.
When the two models use different tokenizations, the same raw text can be segmented differently, so a student-generated token may not correspond to a natural token under the teacher~\citep{boizard2025towards,unlocking_on_policy_distillation,minixhofer2025universal}.
For example, the student may generate \texttt{`<think>'} as \texttt{`<', `think', `>'}, while the teacher expects \texttt{`<th', `ink', `>'}.
Then token \texttt{`<'} receives low probability from the teacher, even though both models produce the same semantic content. Similar mismatches arise for special tokens such as end-of-sequence markers.
In this setting, a one-token comparison confuses semantic disagreement with tokenizer mismatch.
Since supervision is applied on a single token, such mismatch can distort the reward signal.

These observations motivate moving beyond one-token supervision: instead of comparing only the sampled token, we compare teacher and student over a set of teacher-supported next-token continuations at each prefix, while retaining token-level updates for stability.

\begin{figure}[t]
    \centering
    \includegraphics[width=0.75\linewidth]{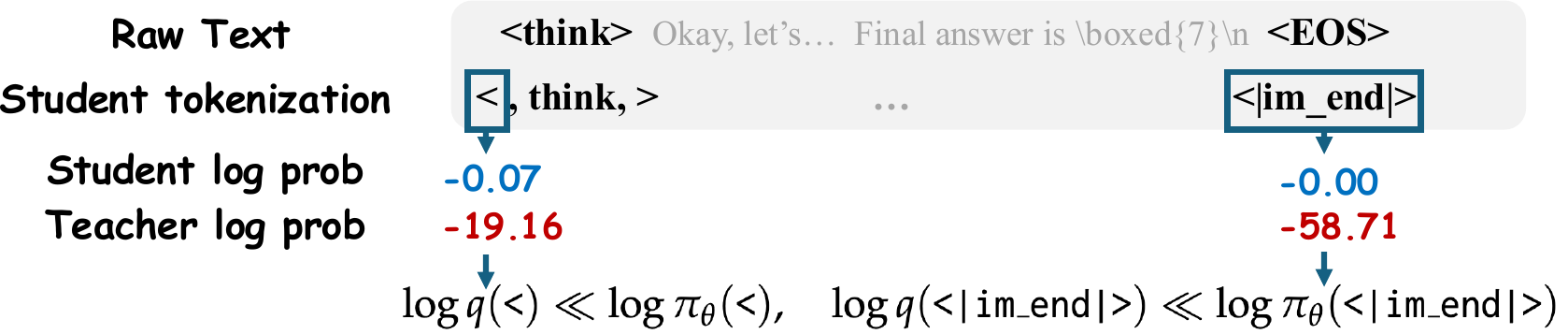}
    \caption{Token-level comparison can penalize semantically correct outputs due to tokenizer mismatch between the teacher and student.}
    \label{fig:mismatch}
\end{figure}

\section{Method}
\label{sec:method}
Our method is designed as a direct response to the failure modes above: it retains token-level OPD, but replaces one-token supervision with a distribution-level comparison over a teacher-selected support set at each prefix.
This yields a truncated reverse-KL objective that preserves the efficiency of local updates while reducing dependence on any single sampled token.
Section~\ref{subsec:topk_method} introduces the objective, and Section~\ref{subsec:practical} describes the practical choices that ensure stable training.
\subsection{Teacher top-\texorpdfstring{$K$}{K} local support matching}
\label{subsec:topk_method}
Instead of comparing teacher and student on a single sampled token, we compare their next-token distributions over a teacher-defined local support set. A natural starting point is the full-vocabulary reverse-KL at prefix $c_t$:
\begin{equation}
\mathcal{L}_{\mathrm{full}}(c_t)
=
\sum_{v \in \mathcal{V}}
\pi_\theta(v \mid c_t)
\log \frac{\pi_\theta(v \mid c_t)}{q(v \mid c_t)}.
\label{eq:full_local_rkl}
\end{equation}
Sampled-token OPD can be viewed as a Monte Carlo approximation to this quantity:
\begin{equation}
\mathcal{L}_{\mathrm{sample}}(c_t, y_t)
=
\log \frac{\pi_\theta(y_t \mid c_t)}{q(y_t \mid c_t)},
\qquad
y_t \sim \pi_\theta(\cdot \mid c_t).
\label{eq:sample_local_rkl}
\end{equation}
This approximation is computationally attractive, but it concentrates the entire update on a single sampled token. We instead compare teacher and student over a teacher-supported candidate set at each prefix.

For each prompt $x$, we sample a group of outputs $\{o_i\}_{i=1}^{G}$ using the student inference policy. Let $c_{i,t} = (x, y_{i,<t})$ be the prefix at position $t$ of output $o_i$, and define the teacher support set
\begin{equation}
S(c_{i,t}) = \mathrm{TopK}_q(c_{i,t}),
\label{eq:topk_support}
\end{equation}
which contains the $K$ highest-probability tokens under the teacher at that prefix.

Within this support, we renormalize both teacher and student distributions:
\begin{equation}
\hat \pi_\theta(v \mid c_{i,t})
=
\frac{\pi_\theta(v \mid c_{i,t})}{\sum_{u \in S(c_{i,t})}\pi_\theta(u \mid c_{i,t})},
\qquad
\hat q(v \mid c_{i,t})
=
\frac{q(v \mid c_{i,t})}{\sum_{u \in S(c_{i,t})}q(u \mid c_{i,t})}.
\label{eq:renorm}
\end{equation}
Our local support matching (LSM) objective averages the truncated reverse-KL over all rollout positions:
\begin{equation}
\mathcal{L}_{\mathrm{LSM}}
=
\mathbb{E}_{x,\, \{o_i\} \sim \pi_{\theta,\mathrm{infer}}}
\left[
\frac{1}{\sum_{i=1}^{G} |o_i|}
\sum_{i=1}^{G}
\sum_{t=1}^{|o_i|}
\sum_{v \in S(c_{i,t})}
\hat \pi_\theta(v \mid c_{i,t})
\log
\frac{\hat \pi_\theta(v \mid c_{i,t})}{\hat q(v \mid c_{i,t})}
\right].
\label{eq:lsm}
\end{equation}
Relative to sampled-token OPD, this replaces a one-token point estimate with a distribution-level comparison over teacher-supported candidates at the same prefix. The update is therefore no longer determined by the sign and magnitude of a single sampled-token log-ratio, while remaining much cheaper than full-vocabulary KL.

\subsection{Practical stabilization choices}
\label{subsec:practical}

\paragraph{Support-set renormalization.}
Renormalization is necessary because the objective is evaluated on a truncated support rather than the full vocabulary. Specifically, we apply a separate softmax over the logits inside the support set, so gradients do not directly propagate to tokens outside this set. Without this step, optimization can become unstable because the teacher and student probability masses inside the support are not directly comparable.

\paragraph{Top-$p$ rollout sampling.}
We generate rollouts with top-$p$ sampling.
Unconstrained sampling occasionally produces very low-probability tokens, creating prefixes on which the teacher signal becomes less informative and optimization less stable.
Top-$p$ sampling keeps trajectories closer to typical continuations and makes the teacher signal more reliable.

\paragraph{Special-token masking.}
We mask problematic special tokens to reduce false negatives caused by incompatible tokenization conventions. This is an orthogonal practical fix: it materially helps sampled-token OPD in our experiments, while our local support objective is much less sensitive to it.
In principle, one could also merge multi-token marker variants or average over equivalent tokenizations, but we do not pursue those tokenizer-specific remedies here because masking is the simplest model-agnostic correction.

\section{Experiments}
\label{sec:experiments}
We evaluate local support matching in three settings: single-task math reasoning (Section~\ref{subsec:single_task}), alternating multi-task training over math and agentic tasks (Section~\ref{subsec:multi_task}), and an additional single-task agentic setting on a smaller student model (Appendix~\ref{app:webshop}). We also present ablations in Section~\ref{subsec:ablations} and provide training-dynamics analysis in Appendix~\ref{app:dynamics}.

\subsection{Setup}
\label{subsec:setup}

We implement local support matching on top of the verl-agent framework~\citep{gigpo2025}, using Qwen2.5-Instruct models as students. We consider two main settings. The first is single-task math reasoning, where OpenThinker3-7B~\citep{openthoughts2025} serves as the teacher and training uses the English portion of DAPO-Math-17K~\citep{yu2025dapo} with a maximum context length of 16K. The second is a multi-task setting that alternates batches between math reasoning and a multi-turn agentic task based on ALFWorld~\citep{alfworld2021}.
In this setting, math uses OpenThinker3-7B~\citep{openthoughts2025} as the teacher, while the agentic side uses the released GiGPO-Qwen2.5-7B-Instruct-ALFWorld checkpoint~\citep{gigpo2025}.

For math reasoning, we report pass@1 on five benchmarks: Math500~\citep{hendrycks2021}, AIME24~\citep{aime24}, AIME25~\citep{aime25}, Minerva~\citep{minerva2022}, and OlympiadBench~\citep{olympiadbench2024}. For ALFWorld~\citep{alfworld2021}, we report success rate by default. In a small number of cases, we additionally report avg@32 on the math benchmarks. More details on experimental setups can be found in Appendix~\ref{app:exp_details}.

\subsection{Single-task math reasoning}
\label{subsec:single_task}

Table~\ref{tab:single_task} shows that local support matching improves over sampled-token OPD in single-task math reasoning. Sampled-token OPD already improves the student from 28.2 to 36.4 average score, but still remains substantially below the teacher. Applying special-token masking to sampled-token OPD further improves the average to 40.7, indicating that tokenizer-related mismatch is a meaningful part of the failure.

Both variants of our method outperform sampled-token OPD and its masked variant in average score. This shows that the gain is not solely attributable to mismatch handling, and instead supports the role of a stronger distribution-level distillation signal. In addition, masking changes our method only modestly (41.7 vs.\ 41.5), consistent with the view that local support matching is less sensitive to tokenizer mismatch than one-token supervision. Additional evidence on WebShop~\citep{webshop2023} is shown in Appendix~\ref{app:webshop}.

\begin{table}[t]
\vspace{-0.6em}
\centering
\small
\resizebox{\linewidth}{!}{
\begin{tabular}{lcccccc}
\toprule
 \textbf{Method} & \textbf{Math500}  & \textbf{AIME24}  & \textbf{AIME25}  & \textbf{Minerva}  & \textbf{OlympiadBench}  & \textbf{Avg.}  \\
\midrule
Qwen2.5-7B-It & 68.2 & 13.3 & 0.0 & 26.5 & 32.9 & 28.2 \\
OpenThinker3-7B & 92.2 & 53.3 & 40.0 & 39.0 & 55.6 & 56.0 \\
Sampled-token OPD & 80.0 & 10.0 & 16.7 & 32.4 & 43.1 & 36.4 \\
Sampled-token OPD w/ mask & 81.4 & \textbf{26.7} & 16.7 & 34.2 & \textbf{44.7} & 40.7 \\
\textbf{Ours w/o mask} & 80.4 & 23.3 & \textbf{26.7} & 34.2 & 43.9 & \textbf{41.7} \\
\textbf{Ours w/ mask} & \textbf{82.0} & 23.3 & 23.3 & \textbf{34.9} & 43.9 & 41.5\\
\bottomrule
\end{tabular}}
\caption{Results on single-task math reasoning.}
\vspace{-1em}
\label{tab:single_task}
\end{table}

\subsection{Multi-task agentic-plus-reasoning training}
\label{subsec:multi_task}

Table~\ref{tab:multi_task} shows that the effect of local support matching differs across the two task families in alternating multi-task training.
The unmasked version raises the average math score from 34.8 to 41.7 (+19.8\%), while maintaining competitive ALFWorld performance. The masked version achieves the best ALFWorld score at 97.7, but gives up part of the math improvement. This pattern suggests that local support matching is especially helpful for the reasoning side of the mixture, where sampled-token signals are more exposed to prefix drift; masking, in contrast, mainly shifts the trade-off toward the agentic task in this run.
Beyond evaluation performance, our objective also yields consistently better optimization dynamics.
We defer the full learning curves and diagnostic plots to Appendix~\ref{app:dynamics}.

\begin{table}[ht]
\centering
\resizebox{\linewidth}{!}{
\begin{tabular}{lccccccc}
\toprule
\multirow{2}{*}[-0.5ex]{\textbf{Method}} & \textbf{Agentic} & \multicolumn{6}{c}{\textbf{Reasoning}} \\
\cmidrule(lr){2-2}\cmidrule(lr){3-8}
 & \textbf{ALFWorld} & \textbf{MATH500} & \textbf{AIME24} & \textbf{AIME25} & \textbf{Minerva} & \textbf{OlympiadBench} & \textbf{Avg.} \\
\midrule
Qwen2.5-7B-It & 21.9 & 68.2 & 13.3 & 0.0 & 26.5 & 32.9 & 28.2 \\
GiGPO-Qwen2.5-7B-It-Alfworld & 95.3 & -- & -- & -- & -- & -- & -- \\
OpenThinker3-7B & -- & 92.2 & 53.3 & 40.0 & 39.0 & 55.6 & 56.0 \\
Sampled-token OPD & 90.6 & 74.8 & 13.3 & 13.3 & 32.1 & 40.5 & 34.8 \\
Sampled-token OPD w/ mask & 93.8 & 76.0 & 20.0 & 13.3 & 33.5 & 40.4 & 36.6 \\
\textbf{Ours w/o mask} & 95.3 & \textbf{82.0} & \textbf{33.3} & \textbf{16.7} & 32.7 & \textbf{44.0} & \textbf{41.7} \\
\textbf{Ours w/ mask} & \textbf{97.7} & 79.0 & 20.0 & 16.7 & \textbf{34.6} & 42.5 & 38.6 \\
\bottomrule
\end{tabular}}
\caption{Results on batch-alternating multi-task training over ALFWorld and math reasoning.}
\vspace{-0.6em}
\label{tab:multi_task}
\end{table}

\subsection{Ablations}
\label{subsec:ablations}

\begin{figure}[t]
\centering
\begin{subfigure}[t]{0.32\linewidth}
\centering
\includegraphics[width=\linewidth]{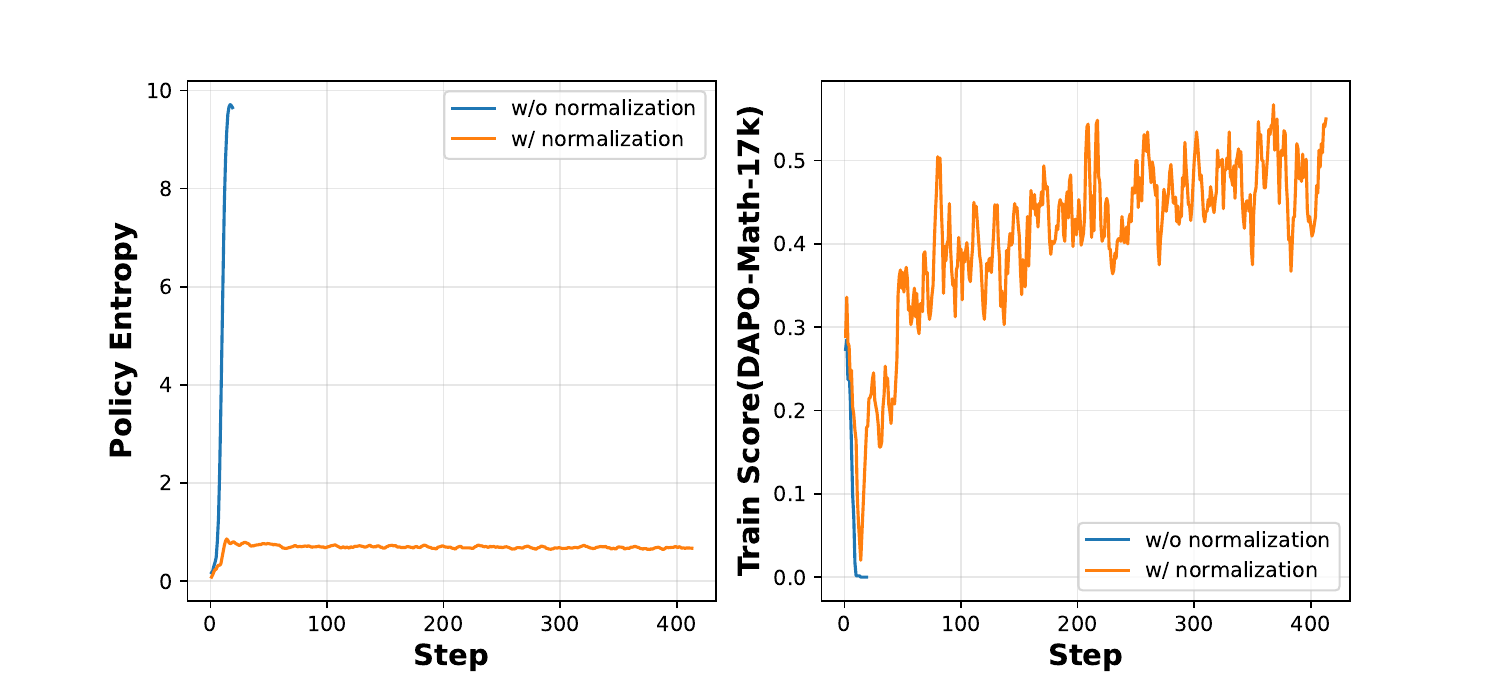}
\caption{Support renormalization.}
\end{subfigure}
\hfill
\begin{subfigure}[t]{0.32\linewidth}
\centering
\includegraphics[width=\linewidth]{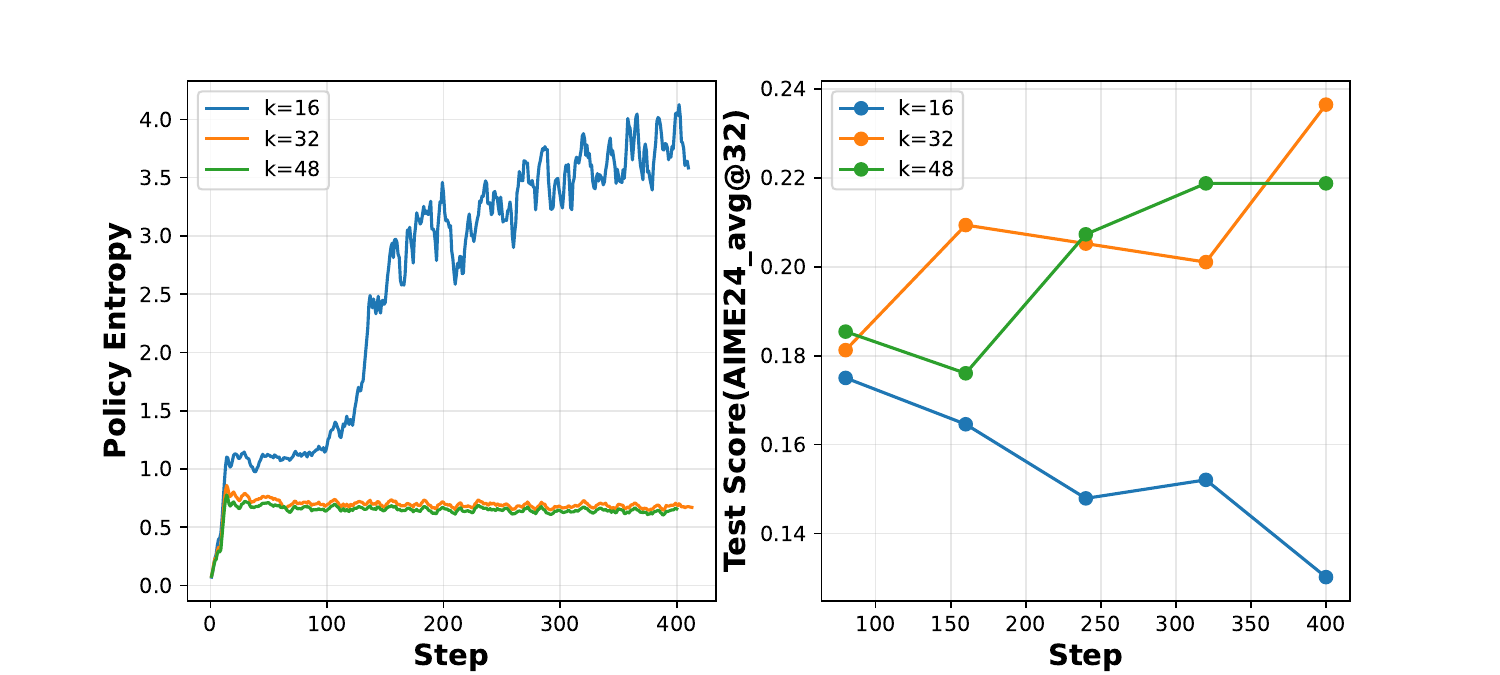}
\caption{Support size $K$.}
\end{subfigure}
\hfill
\begin{subfigure}[t]{0.32\linewidth}
\centering
\includegraphics[width=\linewidth]{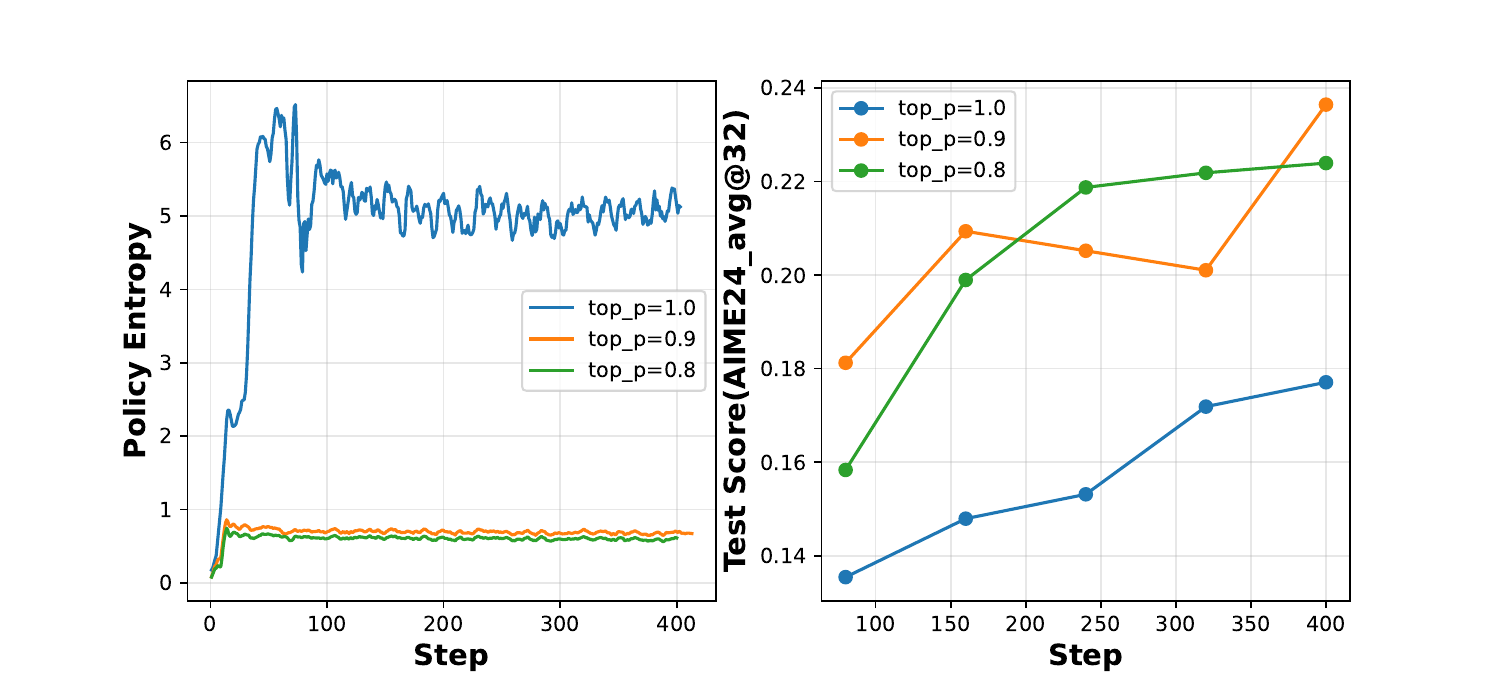}
\caption{Rollout top-$p$.}
\end{subfigure}
\caption{Ablations of the main design choices. Renormalization is required for stability, very small support sets hurt learning, and unconstrained rollout sampling degrades optimization.}
\label{fig:ablations}
\vspace{0.5em}
\end{figure}

Table~\ref{tab:ablation_overview} and Figure~\ref{fig:ablations} suggest that the gains arise from several design choices rather than any single modification. Teacher top-$K$ comparison alone is not sufficient: the rollout policy must also remain in a stable region, and adding top-$p$ sampling turns an initially weaker top-$K$ variant into a stronger configuration. Under the same top-$p$ rollout conditions, teacher top-$K$ local support matching still improves AIME24 avg@32 from 21.6 to 23.6. Renormalization inside the truncated support is essential, as removing it leads to rapid collapse. Performance is not especially sensitive to the exact support size once $K$ is large enough, but training becomes unstable when the support is too small or rollouts are fully unconstrained.

\begin{table}[ht]
\centering
\small
\begin{tabular}{lc}
\toprule
\textbf{Method} & \textbf{AIME24} \textbf{avg@32}  \\
\midrule
Qwen2.5-7B-Instruct & 10.0 \\
OpenThinker3-7B & 63.3 \\
Sampled-token OPD & 20.4 \\
Sampled-token OPD + top-$p$ & 21.6 \\
Teacher top-$K$ & 17.7 \\
Teacher top-$K$ + top-$p$ & 23.6 \\
\bottomrule
\end{tabular}
\caption{Component ablation under single-task math training.}
\vspace{-0.6em}
\label{tab:ablation_overview}
\end{table}

\paragraph{Top-$K$ support variants.}
Our main experiments define the truncated expectation on the teacher's top-$K$ support.
Alternative objectives can compute the expectation over teacher top-$K$ (used in the main results), student top-$K$, or teacher top-$K$ augmented with the student-sampled token. We provide a preliminary comparison in Table~\ref{tab:topk_choice} under both the single-task and multi-task settings.

\begin{table}[t]
\centering

\begin{subtable}{\linewidth}
\centering
\small
\resizebox{\linewidth}{!}{
\begin{tabular}{lccccccc}
\toprule
\multirow{2}{*}[-0.5ex]{\textbf{Method}} & \textbf{avg@32} & \multicolumn{6}{c}{\textbf{pass@1}} \\
\cmidrule(lr){2-2}\cmidrule(lr){3-8}
 & \textbf{AIME24} & \textbf{MATH500} & \textbf{AIME24} & \textbf{AIME25} & \textbf{Minerva} & \textbf{OlympiadBench} & \textbf{Avg.} \\
\midrule
teacher top-$K$ & \textbf{23.6} & 80.4 & 23.3 & \textbf{26.7} & 34.2 & 43.9 & 41.7 \\
student top-$K$ w/ sampled & 22.3 & \textbf{82.4} & \textbf{30.0} & 16.7 & \underline{35.7} & \underline{44.9} & \underline{41.9} \\
teacher top-$K$ w/ sampled & \underline{22.4} & \underline{81.6} & \underline{26.7} & \underline{23.3} & \textbf{36.4} & \textbf{46.7} & \textbf{42.9} \\
\bottomrule
\end{tabular}}
\caption{Alternative KL expectation supports in the single-task setting.}
\label{tab:kl_expectation_single}
\end{subtable}

\vspace{1em}

\begin{subtable}{\linewidth}
\centering
\resizebox{\linewidth}{!}{
\begin{tabular}{lccccccc}
\toprule
\multirow{2}{*}[-0.5ex]{\textbf{Method}} & \textbf{Agentic} & \multicolumn{6}{c}{\textbf{Reasoning}} \\
\cmidrule(lr){2-2}\cmidrule(lr){3-8}
 & \textbf{ALFWorld} & \textbf{MATH500} & \textbf{AIME24} & \textbf{AIME25} & \textbf{Minerva} & \textbf{OlympiadBench} & \textbf{Avg.} \\
\midrule
teacher top-$K$ & \textbf{95.3} & \textbf{82.0} & \textbf{33.3} & \textbf{16.7} & \textbf{32.7} & \textbf{44.0} & \textbf{41.7} \\
student top-$K$ w/ sampled & \textbf{95.3} & \underline{65.6} & \underline{10.0} & \underline{10.0} & \underline{25.0} & \underline{31.6} & \underline{28.4} \\
teacher top-$K$ w/ sampled & \underline{94.5} & 63.2 & \underline{10.0} & \underline{10.0} & 21.0 & 30.1 & 26.9 \\
\bottomrule
\end{tabular}}
\caption{Alternative KL expectation supports in the multi-task setting. We report pass@1 for all math reasoning benchmarks. The final column averages pass@1 metrics only.}
\end{subtable}

\caption{Alternative top-$K$ support variants under single-task and multi-task settings.}
\label{tab:topk_choice}
\vspace{-1em}
\end{table}

In the single-task setting, the three variants achieve broadly comparable results.
The teacher top-$K$ + sampled-token variant attains the highest average pass@1 score, while student top-$K$ performs competitively on several individual benchmarks, and the original teacher top-$K$ gives the best AIME24 avg@32 among the three.
The multi-task results are less uniform.
In that setting, the original teacher top-$K$ variant performs substantially better than the other two alternatives, while both student top-$K$ and teacher top-$K$ + sampled-token variants degrade considerably on the math benchmarks.
We therefore treat these results as evidence that support construction matters, but do not over-interpret the ranking among variants.

Overall, these results suggest that the choice of where the KL expectation is computed can matter in nontrivial ways, especially in the multi-task setting.
We treat this as a partial ablation and a preliminary exploration, and discuss possible causes, including support-set construction and remaining off-policy effects, in Appendix~\ref{app:limitations}.

\section{Conclusion}
\label{sec:conclusion}

This work revisits on-policy distillation (OPD) for LLM post-training from both theoretical and implementation perspectives.
Our theoretical analysis shows why token-level OPD is an attractive approximation for long-horizon training: it is biased relative to sequence-level reverse-KL, but avoids future-reward coupling and has substantially better worst-case variance scaling.
At the same time, our empirical study shows that the standard sampled-token implementation can provide brittle supervision because its signal is imbalanced, can remain misleading on student-drifted prefixes, and is sensitive to tokenizer or special-token mismatch.
Teacher top-$K$ local support matching addresses these issues by preserving local token-level updates while replacing one-token supervision with a truncated distribution-level comparison.
Across single-task math reasoning and alternating agentic-plus-reasoning training, this simple modification improves optimization stability and downstream performance over sampled-token OPD, while also clarifying where teacher matching remains an imperfect proxy for task success.

\newpage
\bibliography{colm2026_conference}
\bibliographystyle{colm2026_conference}

\newpage
\appendix

\renewcommand{\thefigure}{A\arabic{figure}}
\renewcommand{\thetable}{A\arabic{table}}

\setcounter{figure}{0}
\setcounter{table}{0}

\renewcommand{\theequation}{A\arabic{equation}}
\setcounter{equation}{0}

\textbf{\LARGE{Appendix}}
\vspace{5pt}

\section{Discussion and Limitations}
\label{app:limitations}
\paragraph{Sampled-token augmentation of the teacher top-$K$ support.} In our current implementation, the support is $S(c_t)=\mathrm{TopK}_q(c_t)$: teacher and student distributions are renormalized on this teacher-selected set before computing the KL term. However, this is not the only possible design. One can instead use an augmented support $S^+(c_t)=S(c_t)\cup\{y_t\}$, or include the sampled token through an importance-weighted correction that more explicitly preserves unbiasedness with respect to the full-vocabulary reverse-KL~\citep{emapg2026}.
We provide formal definitions of several support-augmentation variants, including alternatives based on teacher- or student-centered supports and EMA-PG-style corrections, and evaluate them in Appendix~\ref{app:topk_variants}. Overall, our default renormalized formulation remains competitive across both single-task and multi-task settings, which is why we adopt it in the main experiments.

\paragraph{Support-set KL is a truncated objective.} Our formulation computes the KL term only over a restricted token subset, i.e., the teacher top-$K$ support, rather than over the full vocabulary. This means that the expected gradient is formed within that subset: tokens outside the support do not receive gradient contributions from this objective. Relative to full-vocabulary reverse-KL, this introduces bias; more generally, it changes where the expectation is taken and which parts of the vocabulary participate in the update.
We describe this here as a property of the estimator, rather than as a settled benefit or drawback.

\paragraph{Training--inference mismatch.} Prefixes are generated under a rollout policy, e.g., top-$p$ sampling in the vLLM engine, while the training engine updates the model without correcting for this sampling process. This may introduce a training--inference mismatch~\citep{liu-li-2025-rl-collapse,yao2025offpolicy}, which we leave to ongoing work.

\paragraph{Teacher matching remains an imperfect proxy for task success.}
Even when OPD is well defined as a teacher-matching objective, the resulting reward can still diverge from the underlying notion of successful behavior. Our reward-hacking cases make this gap concrete: locally teacher-preferred continuations can remain rewardable even when the overall trajectory is already unhelpful or harmful. A noticeable gap to the teacher also remains in our experiments, which suggests that better local supervision is only one part of the distillation problem, especially when teacher and student differ substantially. Closing that gap may require stronger rollout control, better handling of distribution shift, better use of teacher uncertainty, and combinations with outcome-verifiable rewards.

\section{Future Directions}
\label{app:future_directions}

\paragraph{Top-$p$ truncation as an adaptive support.} Another route toward a more compute-efficient reverse-KL objective is top-$p$ truncation instead of top-$K$ truncation, where the KL divergence is computed on the subset of tokens whose cumulative probability mass reaches a prescribed top-$p$ threshold.

\paragraph{OPD versus RL in multi-task settings.}
Our multi-task results motivate a more direct comparison between OPD and RL as transfer mechanisms. In RL, positive or negative transfer can be read directly from environment reward across tasks. In OPD, the optimization target remains teacher-derived, so transfer is filtered through what the teacher regards as locally preferable behavior. This distinction may help explain why our multi-task gains are strongest on the math side and why nearby support-set definitions become less uniform in that setting. A matched-task, matched-compute comparison between OPD and RL would help clarify when teacher-guided transfer tracks environment-level generalization and when the teacher--reward gap becomes the bottleneck.

\paragraph{Continual learning as a testbed.}
Continual learning is another natural setting for OPD. A teacher-guided on-policy objective could act as a retention mechanism while the student adapts to new tasks, but that regime would also stress exactly the issues surfaced in this paper: distribution shift, teacher staleness, and the accumulation of approximation error over long adaptation horizons. Testing OPD there would therefore probe not only whether local support matching mitigates forgetting, but also whether teacher-based objectives remain useful once the student keeps moving away from the teacher's original domain.

\paragraph{Relation to other stabilization directions.}
This work is complementary to directions such as reward-hacking mitigation, EMA-anchor stabilization with top-$K$ KL~\citep{emapg2026}, perturbation-based off-policy correction~\citep{alp2026}, and logit-level fusion between teacher and student rollouts~\citep{logit_fusion_2026}. These methods address different parts of the same broader problem: how to keep teacher-derived signals useful once teacher and student policies begin to diverge~\citep{li2026rethinking}. We view our method as one component in that larger toolbox, rather than as a replacement for those stabilization strategies.

\section{Related Work}
\label{app:related_work}
\paragraph{On-policy distillation.}
Many widely used post-training methods for language models rely on off-policy supervision from fixed datasets, including supervised fine-tuning on demonstration data~\citep{ouyang2022training,openthoughts2025} and conventional knowledge distillation on teacher-provided targets~\citep{hinton2015distilling,sanh2020distilbert}. In autoregressive generation, however, such fixed-prefix training creates a mismatch between the contexts seen during training and the prefixes induced by the student's own test-time rollouts, motivating subsequent work on on-policy distillation (OPD)~\citep{opd2023}, where supervision is computed on student-generated trajectories. \citet{opd2023} formulate generalized knowledge distillation for autoregressive models, while \citet{minillm2023} derive an effective on-policy optimization procedure. More recently, OPD and closely related variants have appeared in large-scale post-training pipelines, including Qwen3~\citep{qwen3_2025}, Thinking Machines Lab~\citep{tmlopd2025}, MiMo-V2-Flash~\citep{mimo_v2_flash_2026}, and GLM-5~\citep{glm5_2026}, underscoring the practical value of on-policy supervision for reasoning transfer, continual adaptation, and capability recovery across training stages. Subsequent work broadens this family in several directions: \citet{ye2026blackboxopd}, \citet{yang2026gopd}, and \citet{ko2026reopold} extend OPD to black-box teachers, more flexible reward formulations, and relaxed imitation objectives, respectively. \citet{ye2026opcd} further connect OPD with context distillation, showing that on-policy supervision can also be used to internalize context- or prompt-conditioned behaviors. Related self- or hindsight-distillation approaches replace the external teacher with guidance derived from privileged information or feedback~\citep{zhao2026opsd,hubotter2026sdpo,wang2026openclawrl,yang2026self}. In contrast, our work does not introduce a new supervision regime; instead, we revisit the standard sampled-token estimator itself and improve its token-level learning signal through teacher top-$K$ local support matching.

\paragraph{KL computation for LLM post-training.}

Sampled KL estimators in LLM post-training are commonly based on forms such as $K_1$, $K_2$, and $K_3$~\citep{schulman2020kl}. These estimators are lightweight, but they retain only the sampled token and therefore discard the richer next-token distribution available from teacher logits in white-box settings. When teacher logits are available, one can in principle match the full next-token distribution, but materializing full-vocabulary logits is often prohibitively memory-intensive for long responses and large models~\citep{hinton2015distilling,kim2016sequence,opd2023}. At one extreme, \citet{zhao2026opsd} use full-vocabulary logit distillation in a self-distillation setting. To reduce the cost of full-distribution matching, \citet{emapg2026} derive an unbiased top-$K$ KL estimator that combines exact computation on a top-$K$ set with a sampled correction for the tail. Related self-distillation methods such as SDPO and OpenClaw-RL also adopt top-$K$ approximations for efficiency, but without explicitly incorporating the sampled token into the truncated support~\citep{hubotter2026sdpo,wang2026openclawrl}. Our method instead computes a truncated reverse-KL objective on the teacher's local top-$K$ support, with both teacher and student distributions renormalized within the selected set. This yields a simple distribution-level alternative to sampled-token OPD without incurring the cost of full-vocabulary matching.

\paragraph{KL divergence in RL for LLMs.}
The choice of KL divergence can substantially affect distribution matching in on-policy distillation and related post-training methods for language models. \citet{minillm2023} argue that reverse KL is often better suited than the forward-KL objective commonly used in conventional distillation because it discourages the student from overestimating the teacher's low-probability regions and tends to yield more precise generation. \citet{opd2023} further show that different divergences induce a quality--diversity trade-off: moving from forward KL toward reverse KL through generalized Jensen--Shannon divergence (JSD) yields increasingly mode-seeking behavior and lower diversity. A related perspective is provided by \citet{chen2025retaining}, who connect the reverse-KL-like, mode-seeking character of on-policy RL to reduced forgetting during post-training. More recently, \citet{jin2026entropyaware} argue that pure reverse-KL-style OPD can become brittle when the teacher distribution has high entropy, and propose augmenting reverse KL with forward KL on high-entropy tokens to better preserve diversity. At the same time, the preferred divergence can depend on the supervision setting. In self-distillation with privileged information, \citet{zhao2026opsd} report that forward KL outperforms reverse KL and JSD, whereas SDPO adopts JSD as a stability-oriented design choice~\citep{hubotter2026sdpo}. In this work, we adopt a reverse-KL formulation, but focus on a different question: how to obtain a stable local reverse-KL-style training signal under sampled-token OPD.

\section{Bias and variance analysis of token-level versus sequence-level OPD}
\label{app:bias_variance}

\subsection{Bias of the token-level estimator}

Recall the sequence-level estimator in causal return-to-go form
\[
\hat g_{\mathrm{seq}} = \sum_{t=1}^{T} \left(\sum_{t'=t}^{T} r_{t'}\right) s_t.
\]
Expanding the inner sum gives
\[
\hat g_{\mathrm{seq}}
=
\sum_{t=1}^{T} r_t s_t
+
\sum_{t=1}^{T}
\sum_{t'=t+1}^{T}
r_{t'} s_t.
\]
Since the token-level estimator keeps only the first term,
\[
\hat g_{\mathrm{tok}} = \sum_{t=1}^{T} r_t s_t,
\]
their expectation gap is
\[
\mathbb{E}[\hat g_{\mathrm{seq}}] - \mathbb{E}[\hat g_{\mathrm{tok}}]
=
\mathbb{E}\left[
\sum_{t=1}^{T}
\sum_{t'=t+1}^{T}
r_{t'} s_t
\right].
\]
This makes explicit that token-level OPD removes the future-reward coupling terms and is therefore generally biased with respect to the sequence-level objective.

\subsection{Worst-case variance upper bounds}

Assume there exist constants $B_r, B_s > 0$ such that
\[
|r_t| \le B_r,
\qquad
\|s_t\| \le B_s
\quad \text{for all } t.
\]
For the token-level estimator,
\[
\|\hat g_{\mathrm{tok}}\|
\le
\sum_{t=1}^{T} |r_t| \, \|s_t\|
\le
T B_r B_s,
\]
which implies
\[
\mathbb{E}\|\hat g_{\mathrm{tok}}\|^2 \le T^2 B_r^2 B_s^2.
\]
Using $\mathrm{Var}(X) \le \mathbb{E}\|X\|^2$, we obtain
\[
\mathrm{Var}(\hat g_{\mathrm{tok}}) = O(T^2).
\]

For the sequence-level estimator, define
\[
R = \sum_{t=1}^{T} r_t,
\qquad
S = \sum_{t=1}^{T} s_t,
\qquad
\hat g_{\mathrm{seq}} = RS.
\]
Then
\[
|R| \le T B_r,
\qquad
\|S\| \le T B_s,
\]
so
\[
\|\hat g_{\mathrm{seq}}\|
\le
T^2 B_r B_s,
\qquad
\mathbb{E}\|\hat g_{\mathrm{seq}}\|^2 \le T^4 B_r^2 B_s^2.
\]
Therefore,
\[
\mathrm{Var}(\hat g_{\mathrm{seq}}) = O(T^4).
\]

\subsection{Discussion}

The sequence-level estimator is closer to the exact trajectory-level objective, but it couples each score term with many future rewards. In worst-case scaling, this changes variance growth from quadratic to quartic in sequence length. The argument is conservative, but it captures why stronger reward coupling can become problematic in long-horizon training.

\section{Toy experiment details}
\label{app:toy}

\subsection{Environment}

We use a two-task one-dimensional continuous-control environment to visualize how stronger return coupling changes OPD optimization. The student policy is a three-layer MLP with roughly 4K parameters. Its input is a three-dimensional vector containing task identity, current position, and normalized time step. The policy outputs the mean and standard deviation of a Gaussian action distribution, and the state transition is
\[
s_{t+1} = s_t + \delta,
\qquad
\delta \sim \mathcal{N}(\mu, \sigma).
\]
The two tasks are mirror images of each other: the left task starts from $+2$ and targets $-3$, while the right task starts from $-2$ and targets $+3$. We first train separate teachers with REINFORCE and then distill them into a shared student with alternating-task OPD.

\subsection{Gradient variance estimation}

At each training step, we split a batch of $B=64$ trajectories into $M=8$ micro-batches. For each micro-batch $m$, we compute a loss $\mathcal{L}_m$ and the corresponding gradient vector $\mathbf g_m$ on the output layer parameters. We then estimate gradient variance by
\[
\mathrm{Var}(\mathbf g)
=
\frac{1}{M}
\sum_{m=1}^{M}
\left\|
\mathbf g_m - \bar{\mathbf g}
\right\|^2,
\qquad
\bar{\mathbf g}
=
\frac{1}{M}
\sum_{m=1}^{M}
\mathbf g_m.
\]
We use this quantity only as a qualitative proxy, but it is sufficient for comparing relative variance across different $\gamma$ settings.

\subsection{Additional Results of Toy Experiments}
\begin{figure}[H]
    \centering
    \begin{subfigure}{\linewidth}
        \centering
        \includegraphics[width=0.75\linewidth]{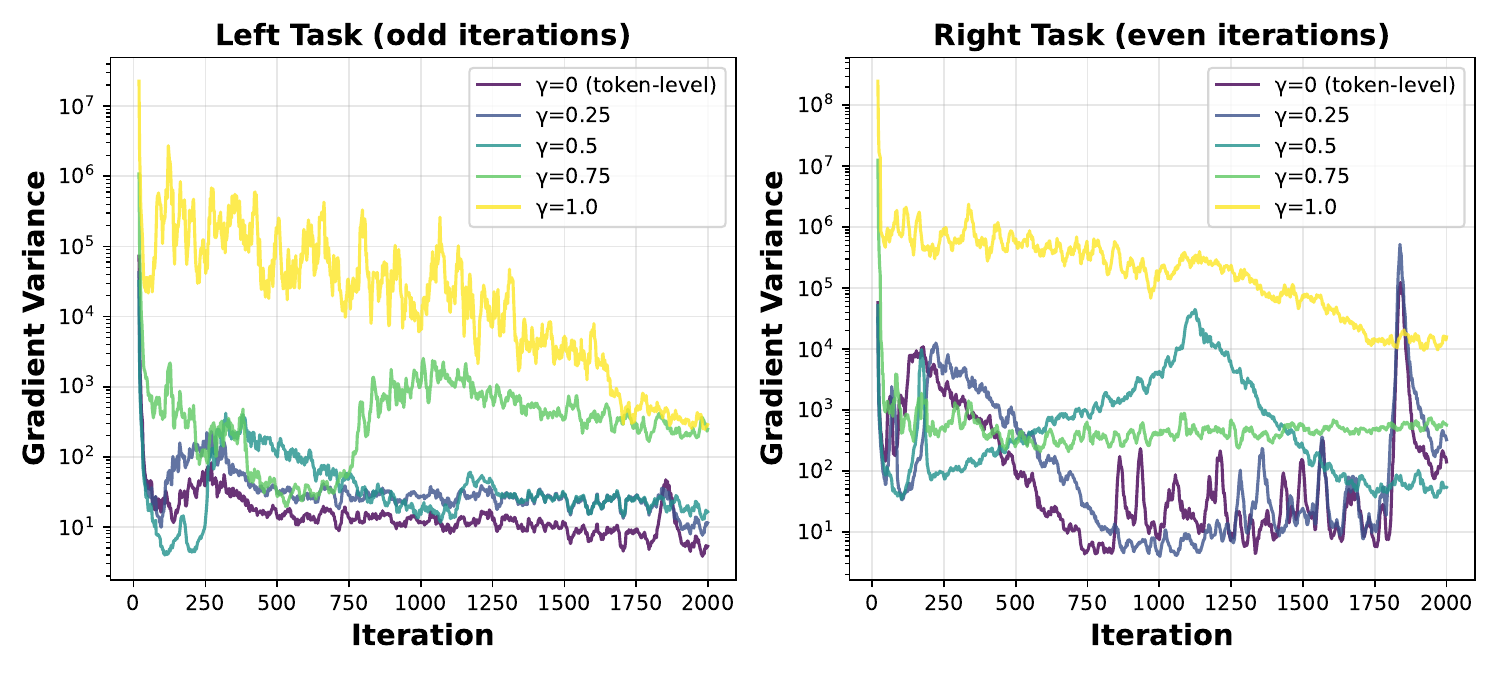}
    \end{subfigure}
    \vspace{0.5em}
    \begin{subfigure}{\linewidth}
        \centering
        \includegraphics[width=0.75\linewidth]{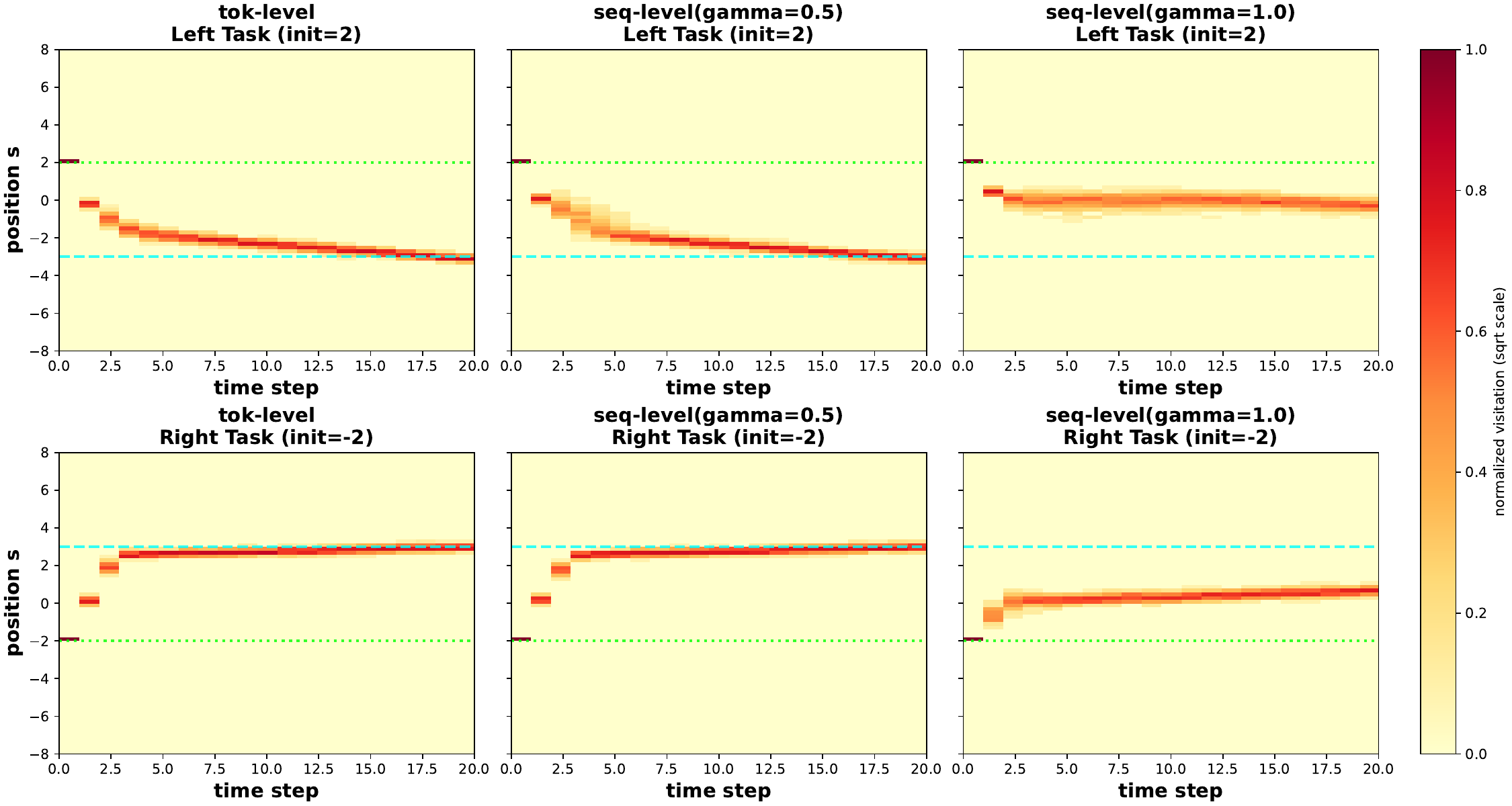}
    \end{subfigure}
    \caption{Toy experiment with random seed 42: gradient variance and state visitation.}
    \label{app:toy_42}
\end{figure}

\begin{figure}[H]
    \centering
    \begin{subfigure}{\linewidth}
        \centering
        \includegraphics[width=0.75\linewidth]{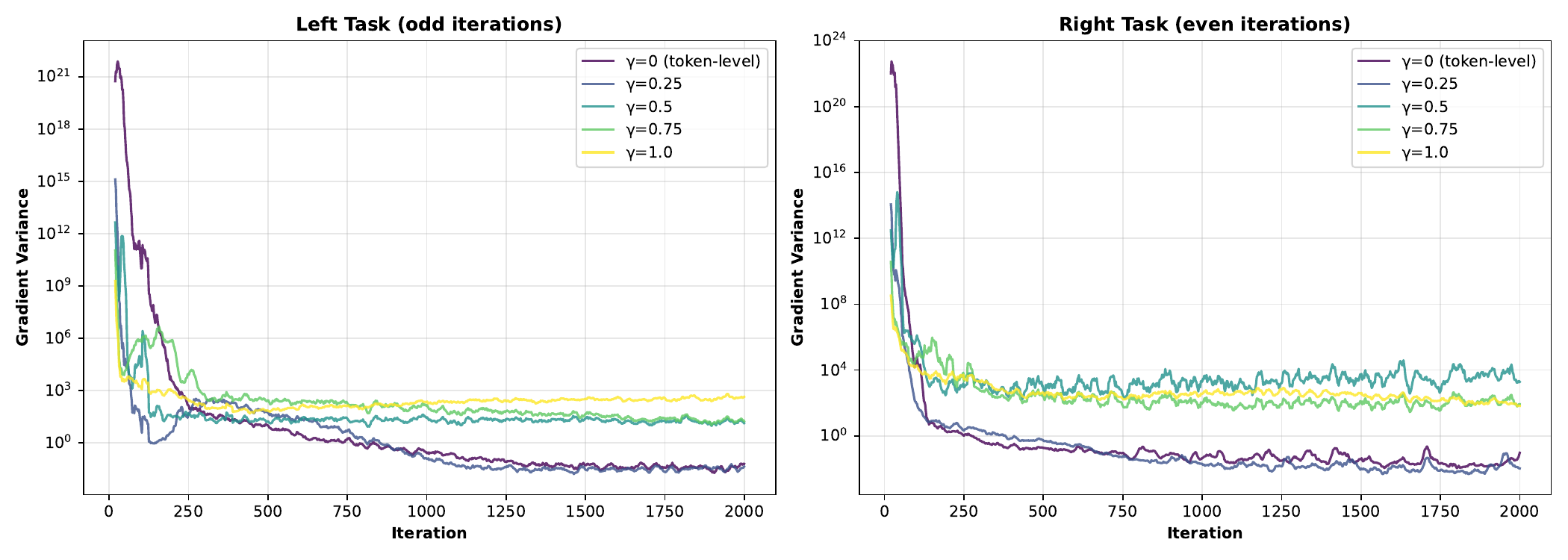}
    \end{subfigure}
    \vspace{0.5em}
    \begin{subfigure}{\linewidth}
        \centering
        \includegraphics[width=0.75\linewidth]{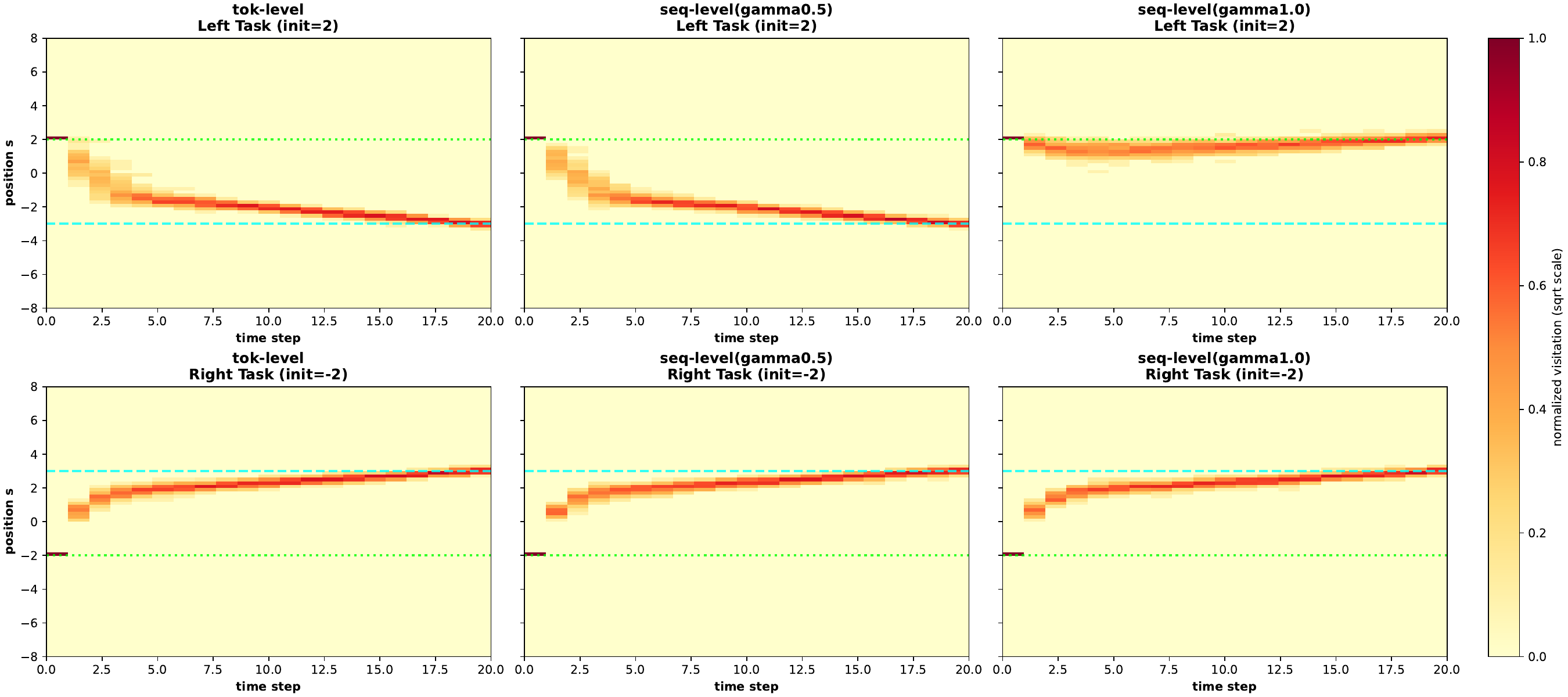}
    \end{subfigure}
    \caption{Toy experiment with random seed 43: gradient variance and state visitation.}
    \label{app:toy_43}
\end{figure}

\begin{figure}[H]
    \centering
    \begin{subfigure}{\linewidth}
        \centering
        \includegraphics[width=0.75\linewidth]{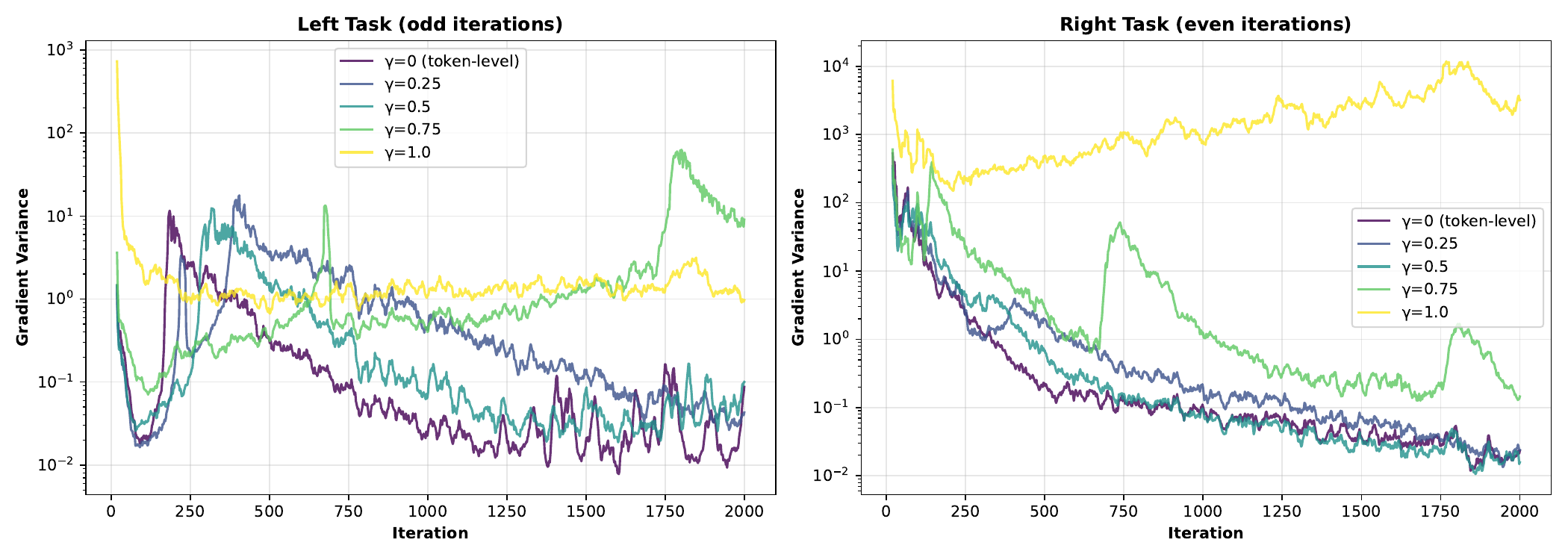}
    \end{subfigure}
    \vspace{0.5em}
    \begin{subfigure}{\linewidth}
        \centering
        \includegraphics[width=0.75\linewidth]{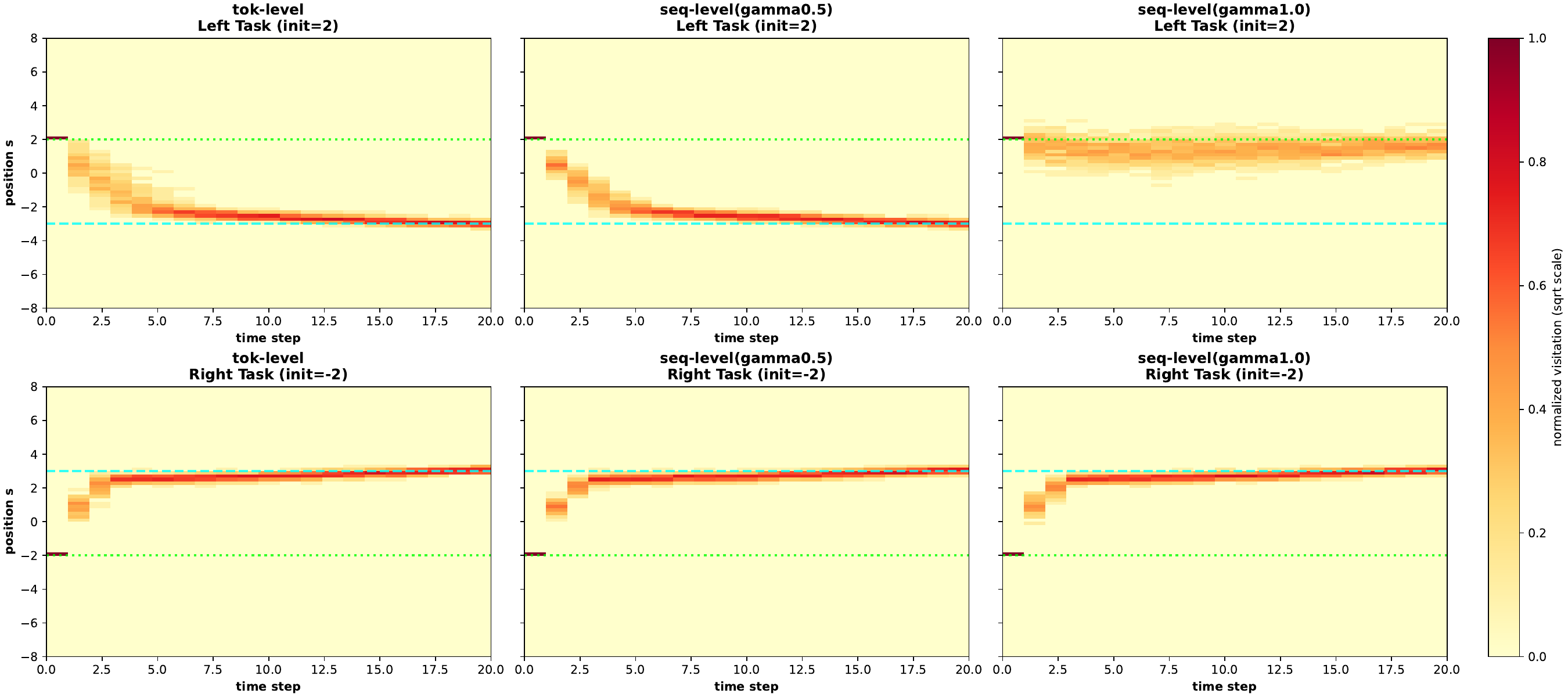}
    \end{subfigure}
    \caption{Toy experiment with random seed 2026: gradient variance and state visitation.}
    \label{app:toy_2026}
\end{figure}

Figures~\ref{app:toy_42}, \ref{app:toy_43}, and \ref{app:toy_2026} report gradient-variance curves and corresponding state-visitation heatmaps for different OPD estimators ($\gamma \in \{0.0, 0.25, 0.5, 0.75, 1.0\}$) across three random seeds. Although the exact magnitudes vary by seed, the qualitative pattern is consistent. All settings exhibit large variance spikes during early optimization, and larger $\gamma$ typically remains at a higher variance level later in training. In several runs, the variance under $\gamma=0.75$ or $\gamma=1.0$ stays one to several orders of magnitude above that of smaller $\gamma$ values.
Across runs, token-level OPD ($\gamma=0$) consistently learns trajectories that move toward the target states for both tasks. Intermediate values of $\gamma$ remain qualitatively similar but become more diffuse. When $\gamma$ approaches the sequence-level case ($\gamma=1.0$), the learned trajectories often deviate from the desired direction and stabilize around sub-optimal regions.

\section{Experiment Setups}
\label{app:exp_details}
We build our training pipeline on verl\footnote{\url{https://github.com/verl-project/verl}}~\citep{verl2025} and the verl-agent framework\footnote{\url{https://github.com/langfengQ/verl-agent}}~\citep{gigpo2025}, and conduct all experiments on a node with 8 NVIDIA H100 GPUs. We report the settings and hyperparameters used in our experiments in Table~\ref{tab:exp_settings}.
\begin{table}[H]
\centering
\renewcommand{\arraystretch}{1.3}
\resizebox{\textwidth}{!}{
\begin{tabular}{llccc}
\toprule
\textbf{Category} & \textbf{Item} & \textbf{Math ($\S$~\ref{subsec:single_task},$\S$~\ref{subsec:multi_task})} & \textbf{ALFWorld ($\S$~\ref{subsec:multi_task})} & \textbf{WebShop ($\S$~\ref{app:webshop})} \\
\midrule
\multirow{2}{*}{Model}
& Student
& Qwen2.5-7B-Instruct
& Qwen2.5-7B-Instruct
& Qwen2.5-1.5B-Instruct \\
& Teacher
& OpenThinker3-7B
& GiGPO-Qwen2.5-7B-It-ALFWorld
& GiGPO-Qwen2.5-1.5B-It-WebShop \\
\midrule
\multirow{3}{*}{Environment}
& Max prompt length
& 2048
& 2048
& 2048 \\
& Max response length
& 16384
& 512
& 512 \\
& Max turns
& --
& 30
& 15 \\
\midrule
\multirow{3}{*}{Rollout}
& Group size
& 8
& 8
& 8 \\
& Top-$p$*
& 0.9
& 0.9
& 0.9 \\
& Temperature
& 1
& 1
& 1 \\
\midrule
\multirow{2}{*}{Evaluation}
& Top-$p$
& 0.9
& 0.9
& 0.9 \\
& Temperature
& 1
& 1
& 1 \\
\midrule
\multirow{7}{*}{Training}
& Teacher top-$K$
& 32
& 32
& 32 \\
& Optimizer
& AdamW
& AdamW
& AdamW \\
& Total training steps$^\dagger$
& 400
& 400
& 60 \\
& Learning rate
& $2\times10^{-6}$
& $2\times10^{-6}$
& $2\times10^{-6}$ \\
& Warmup steps
& 0
& 0
& 0 \\
& Batch size
& 128
& 128
& 128 \\
& Mini batch size
& 64
& 64
& 64 \\
\bottomrule
\end{tabular}
}
\caption{Training and evaluation settings. * Rollout top-$p$ is used in local support matching, but not in the sampled-token OPD baseline. $^\dagger$ Training steps denote the total number of training updates; in the alternating multi-task setting, 400 total steps correspond to 200 math updates and 200 ALFWorld updates.}
\label{tab:exp_settings}
\end{table}

\section{Additional Results}

\subsection{Additional Evidence on Single-task WebShop}
\label{app:webshop}

To test whether the same trend transfers beyond the main math and ALFWorld settings, we additionally evaluate single-task WebShop~\citep{webshop2023} using Qwen2.5-1.5B-Instruct as the student. As shown in Table~\ref{tab:webshop}, local support matching improves over sampled-token OPD on both task score and success rate, increasing success rate from 50.0 to 57.8. Although the gap to the teacher remains substantial, the result suggests that the method transfers beyond the main math/agentic settings. We do not include a masking variant in this setting because the teacher is obtained by RL training from the same base model, so tokenizer or special-token mismatch is less likely to be a dominant factor here.

\begin{table}[ht]
    \centering
    \small
    \begin{tabular}{lcc}
    \toprule
         \textbf{Method} & \textbf{Task score} & \textbf{Success rate} \\
    \midrule
    Qwen2.5-1.5B-It & 12.7 & 2.3 \\
    GiGPO-Qwen2.5-1.5B-Webshop~\citep{gigpo2025} & 81.9 & 66.4 \\
    Sampled-token OPD & 73.0 & 50.0 \\
    \textbf{Ours} & \textbf{75.1} & \textbf{57.8} \\
    \bottomrule
    \end{tabular}
    \caption{Results on single-task WebShop with Qwen2.5-1.5B-Instruct as the student. Local support matching improves over sampled-token OPD on both task score and success rate, reducing the gap to the teacher. We do not include a masking variant here because the teacher is obtained by RL training from the same base model, making tokenizer-mismatch effects less central in this setting.}
    \label{tab:webshop}
\end{table}

\subsection{Training Dynamics and Alignment}
\label{app:dynamics}

\begin{figure}[ht]
\centering
\includegraphics[width=0.6\linewidth]{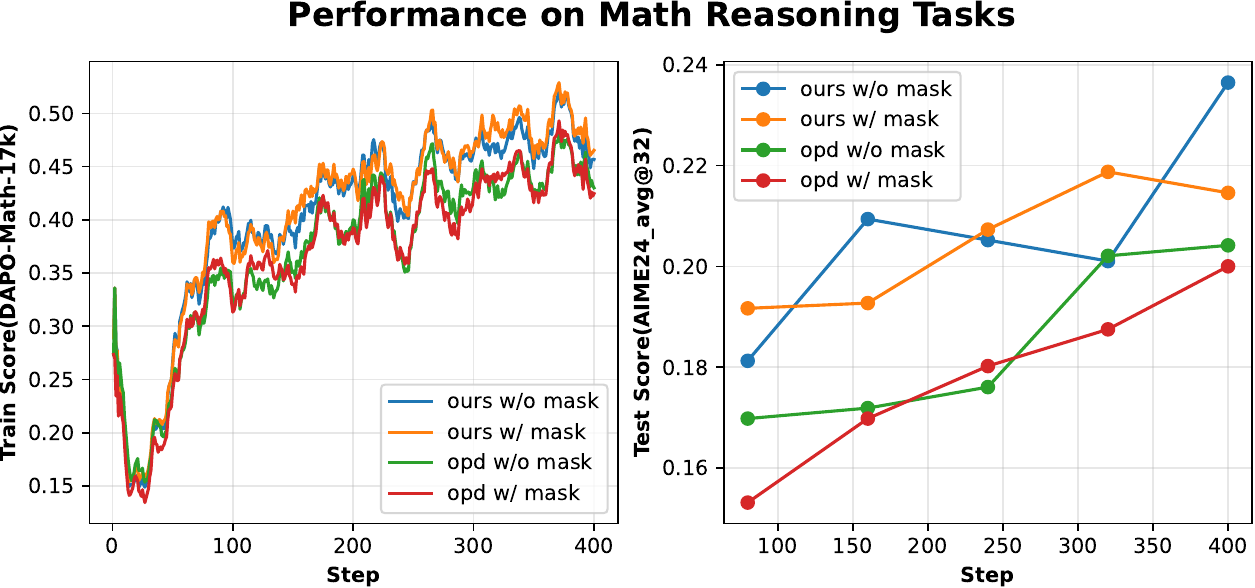}
\caption{Single-task training curves for math reasoning. Local support matching improves both training reward and evaluation performance over the course of training.}
\label{fig:learning_curves_single}
\end{figure}

\begin{figure}[ht]
\centering
\includegraphics[width=\linewidth]{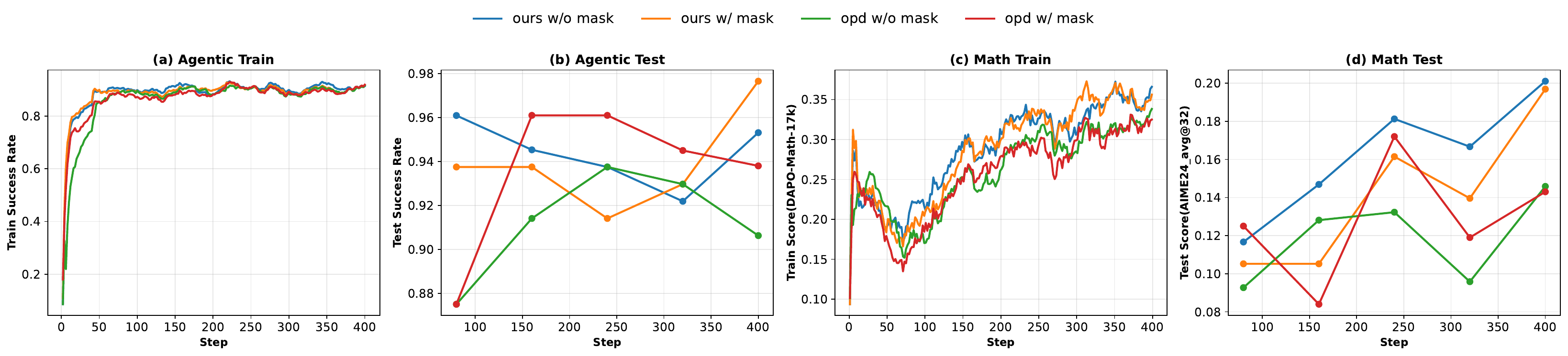}
\caption{Multi-task learning curves for ALFWorld and math reasoning. The main gains appear on the math side while agentic performance remains strong.}
\label{fig:learning_curves_multi}
\end{figure}

\begin{figure}[t]
\centering
\begin{subfigure}[t]{0.49\linewidth}
\centering
\includegraphics[width=\linewidth]{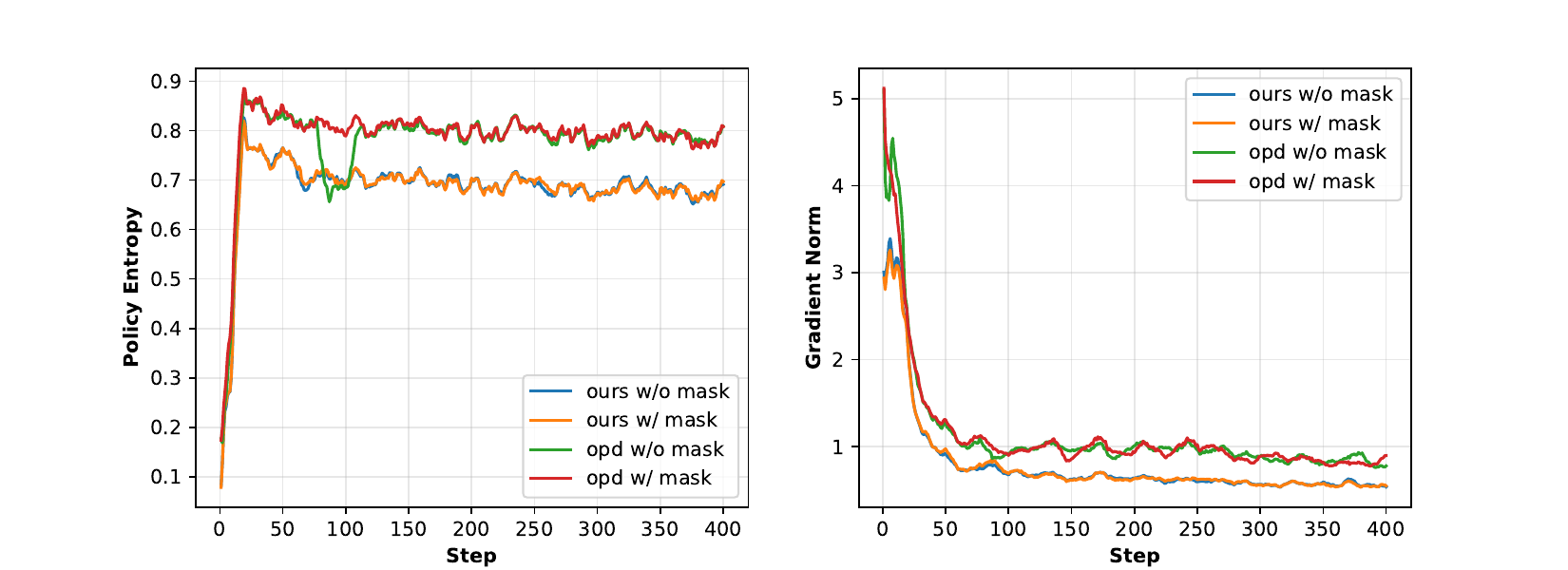}
\caption{Single-task optimization statistics.}
\end{subfigure}
\hfill
\begin{subfigure}[t]{0.49\linewidth}
\centering
\includegraphics[width=\linewidth]{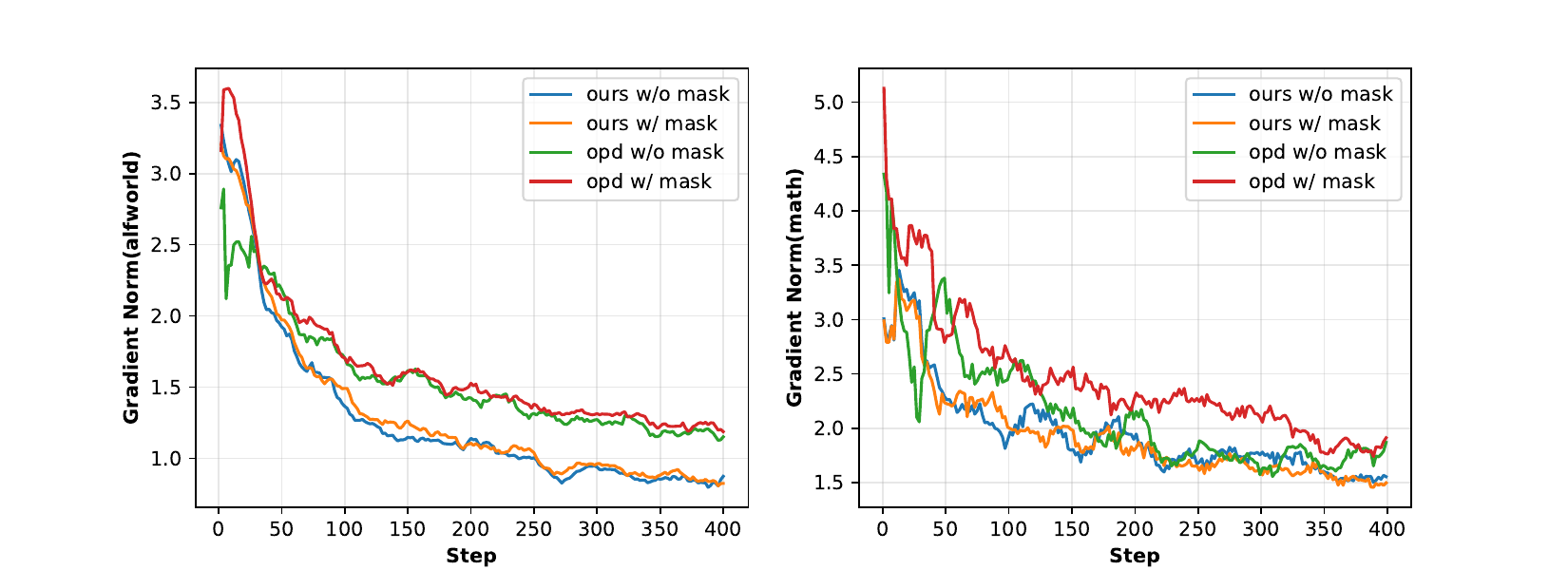}
\caption{Multi-task gradient norms.}
\end{subfigure}

\vspace{0.5em}

\begin{subfigure}[t]{0.49\linewidth}
\centering
\includegraphics[width=\linewidth]{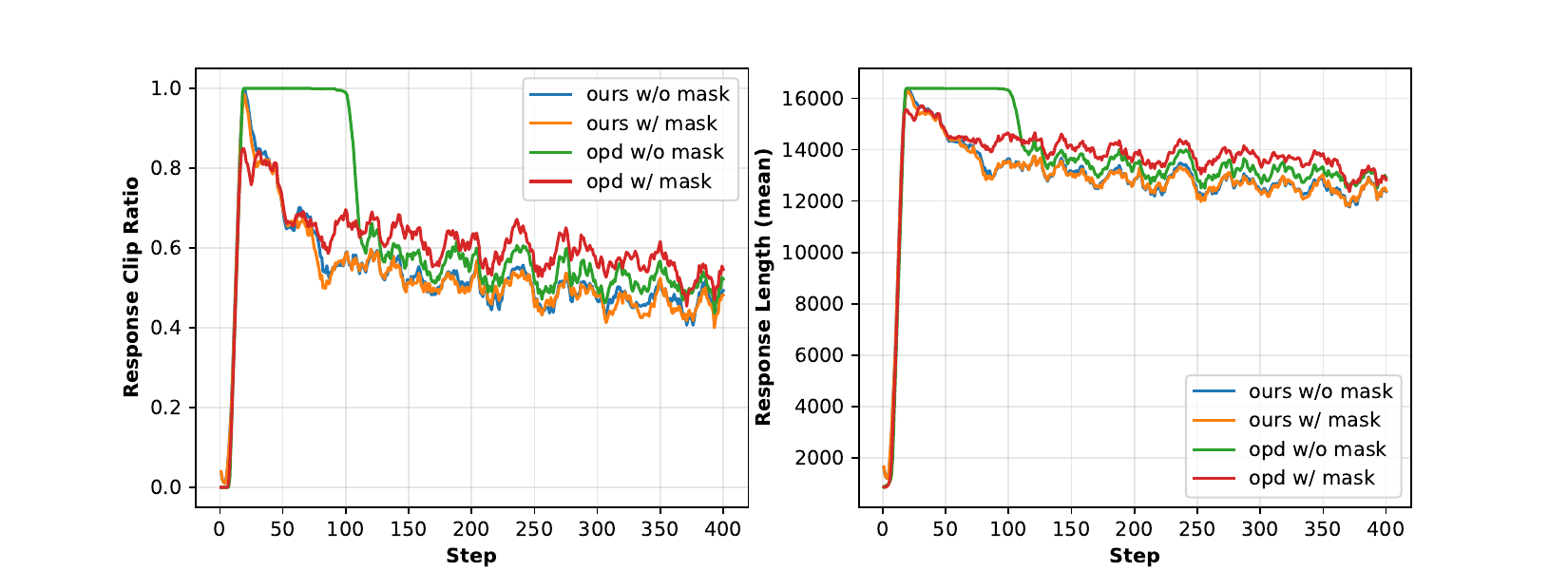}
\caption{Response length statistics.}
\end{subfigure}
\begin{subfigure}[t]{0.49\linewidth}
\centering
\includegraphics[width=\linewidth]{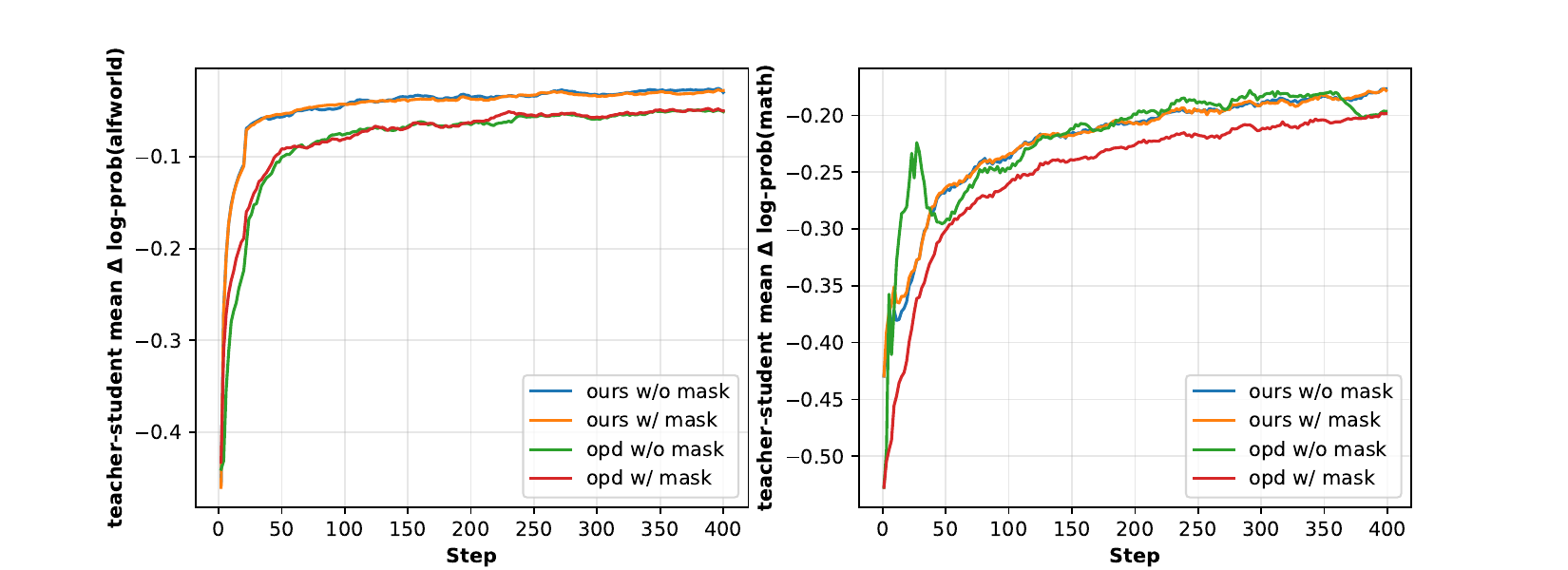}
\caption{Teacher--student log-probability gaps.}
\end{subfigure}
\caption{Optimization and alignment diagnostics. Relative to sampled-token OPD, local support matching yields smaller gradient norms, fewer clipping-boundary hits, shorter responses, and teacher--student log-probability gaps closer to zero.}
\label{fig:optimization}
\end{figure}

Figures~\ref{fig:learning_curves_single}, \ref{fig:learning_curves_multi}, and \ref{fig:optimization} provide a more detailed view of the optimization dynamics.

\paragraph{Better learning curves.}
On math reasoning, our method improves both training reward and evaluation performance across most of training rather than only at the final checkpoint. This pattern holds in both the single-task setting and the alternating multi-task setting.

\paragraph{More stable optimization.}
Our method yields smaller gradient norms and lower clipping-boundary fractions while maintaining sufficient policy entropy, indicating more stable optimization. We also observe that special-token masking substantially reduces the clipping-boundary fraction of sampled-token OPD during early and middle training, while having only minor effects on our method.

\paragraph{Improved teacher--student alignment.}
The teacher--student log-probability gap on sampled tokens also moves closer to zero, suggesting that the truncated local support objective improves alignment even under the sampled-token diagnostic used by the baseline.

\subsection{Additional Results of Top-\texorpdfstring{$K$}{K} Variants}
\label{app:topk_variants}
We primarily evaluate five top-$K$ variants in our experiments. Below, we present their formulations under unified notation and evaluate them in both single-task math and multi-task settings. All experimental setups are kept the same as in the main experiments.
\subsubsection{Formulations of Five Top-\texorpdfstring{$K$}{K} Variants}
\paragraph{Shared notation.}
For each prompt $x$, we sample a group of outputs $\{o_i\}_{i=1}^G$ from the student inference policy $\pi_{\mathrm{infer}}$. Let $y_{i,t}$ denote the token at position $t$ in output $o_i$, and let
\[
c_{i,t} = (x, y_{i,<t})
\]
be the corresponding prefix. We write
\[
\mathcal T_K^\pi(c_{i,t}) = \operatorname{TopK}_\pi(c_{i,t}),
\qquad
\mathcal T_K^q(c_{i,t}) = \operatorname{TopK}_q(c_{i,t}),
\]
for the student and teacher top-$K$ token sets at prefix $c_{i,t}$.

All variants below share the same rollout-level aggregation:
\[
\mathcal L[\ell]
=
\mathbb E_{x,\{o_i\}\sim\pi_{\mathrm{infer}}}
\left[
\frac{1}{\sum_{i=1}^G |o_i|}
\sum_{i=1}^G \sum_{t=1}^{|o_i|}
\ell(c_{i,t}, y_{i,t})
\right].
\]
They differ only in how the local loss $\ell(c_{i,t}, y_{i,t})$ is defined.

\paragraph{Renormalized local reverse-KL.}
For any support set $\mathcal S(c_{i,t}) \subseteq \mathcal V$, define the renormalized student and teacher distributions on $\mathcal S(c_{i,t})$ by
\[
\bar\pi_\theta(v \mid c_{i,t}; \mathcal S)
=
\frac{\pi_\theta(v \mid c_{i,t})}
{\sum_{u \in \mathcal S(c_{i,t})} \pi_\theta(u \mid c_{i,t})},
\qquad
\bar q(v \mid c_{i,t}; \mathcal S)
=
\frac{q(v \mid c_{i,t})}
{\sum_{u \in \mathcal S(c_{i,t})} q(u \mid c_{i,t})},
\quad v \in \mathcal S(c_{i,t}).
\]
The corresponding local reverse-KL is
\[
\ell_{\mathrm{renorm}}(c_{i,t}; \mathcal S)
=
\sum_{v \in \mathcal S(c_{i,t})}
\bar\pi_\theta(v \mid c_{i,t}; \mathcal S)
\log
\frac{\bar\pi_\theta(v \mid c_{i,t}; \mathcal S)}
{\bar q(v \mid c_{i,t}; \mathcal S)}.
\]

\paragraph{Variant 1: teacher top-$K$ renormalization (our method).}
Our default local support matching (LSM) method uses the teacher top-$K$ support
\[
\mathcal S_{\mathrm{LSM}}(c_{i,t}) = \mathcal T_K^q(c_{i,t}),
\]
and defines
\[
\ell_{\mathrm{LSM}}(c_{i,t}, y_{i,t})
=
\ell_{\mathrm{renorm}}(c_{i,t}; \mathcal S_{\mathrm{LSM}}).
\]
The overall objective is
\[
\mathcal L_{\mathrm{LSM}} = \mathcal L[\ell_{\mathrm{LSM}}].
\]

\paragraph{Variant 2: student top-$K$ + sampled token + renormalization.}
For this variant, the support is obtained by augmenting the student top-$K$ set with the sampled token when needed:
\[
\mathcal S_{\pi+y}(c_{i,t})
=
\begin{cases}
\mathcal T_K^\pi(c_{i,t}), & y_{i,t} \in \mathcal T_K^\pi(c_{i,t}),\\[4pt]
\mathcal T_K^\pi(c_{i,t}) \cup \{y_{i,t}\}, & y_{i,t} \notin \mathcal T_K^\pi(c_{i,t}).
\end{cases}
\]
We then define
\[
\ell_{\pi+y}(c_{i,t}, y_{i,t})
=
\ell_{\mathrm{renorm}}(c_{i,t}; \mathcal S_{\pi+y}),
\]
and
\[
\mathcal L_{\pi+y} = \mathcal L[\ell_{\pi+y}].
\]

\paragraph{Variant 3: teacher top-$K$ + sampled token + renormalization.}
Similarly, this variant augments the teacher top-$K$ set with the sampled token:
\[
\mathcal S_{q+y}(c_{i,t})
=
\begin{cases}
\mathcal T_K^q(c_{i,t}), & y_{i,t} \in \mathcal T_K^q(c_{i,t}),\\[4pt]
\mathcal T_K^q(c_{i,t}) \cup \{y_{i,t}\}, & y_{i,t} \notin \mathcal T_K^q(c_{i,t}).
\end{cases}
\]
Its local and rollout-level losses are
\[
\ell_{q+y}(c_{i,t}, y_{i,t})
=
\ell_{\mathrm{renorm}}(c_{i,t}; \mathcal S_{q+y}),
\qquad
\mathcal L_{q+y} = \mathcal L[\ell_{q+y}].
\]

\paragraph{EMA-PG-style local loss.}
The remaining two variants follow the EMA-PG decomposition~\citep{emapg2026}, where a top-$K$ truncated head term is combined with an importance-weighted sampled-token tail correction to obtain an unbiased estimator of the full-vocabulary reverse-KL. For any support set $\mathcal S(c_{i,t}) \subseteq \mathcal V$, define the truncated head term
\[
\ell_{\mathrm{head}}(c_{i,t}; \mathcal S)
=
\sum_{v \in \mathcal S(c_{i,t})}
\pi_\theta(v \mid c_{i,t})
\log
\frac{\pi_\theta(v \mid c_{i,t})}{q(v \mid c_{i,t})}.
\]
Let $\operatorname{sg}(\cdot)$ denote stop-gradient, and define
\[
r_{i,t}
=
\frac{\pi_\theta(y_{i,t} \mid c_{i,t})}
{\operatorname{sg}\!\left(\pi_\theta(y_{i,t} \mid c_{i,t})\right)},
\qquad
s_{i,t}
=
\operatorname{sg}\!\left(
\frac{\pi_\theta(y_{i,t} \mid c_{i,t})}
{\pi_{\mathrm{infer}}(y_{i,t} \mid c_{i,t})}
\right),
\]
where $s_{i,t}$ is the sampling-policy correction (optionally clipped, as in EMA-PG). The sampled-token tail correction is
\[
\ell_{\mathrm{tail}}(c_{i,t}, y_{i,t}; \mathcal S)
=
s_{i,t}\,
\mathbf 1\!\left[y_{i,t} \notin \mathcal S(c_{i,t})\right]\,
r_{i,t}\,
\operatorname{sg}\!\left(
\log
\frac{\pi_\theta(y_{i,t} \mid c_{i,t})}
{q(y_{i,t} \mid c_{i,t})}
\right).
\]
The corresponding local EMA-PG loss is
\[
\ell_{\mathrm{EMA}}(c_{i,t}, y_{i,t}; \mathcal S)
=
\ell_{\mathrm{head}}(c_{i,t}; \mathcal S)
+
\ell_{\mathrm{tail}}(c_{i,t}, y_{i,t}; \mathcal S).
\]

\paragraph{Variant 4: student top-$K$ + EMA-PG correction.}
This variant uses the student top-$K$ support
\[
\mathcal S_{\pi\text{-EMA}}(c_{i,t}) = \mathcal T_K^\pi(c_{i,t}),
\]
and defines
\[
\ell_{\pi\text{-EMA}}(c_{i,t}, y_{i,t})
=
\ell_{\mathrm{EMA}}(c_{i,t}, y_{i,t}; \mathcal S_{\pi\text{-EMA}}),
\qquad
\mathcal L_{\pi\text{-EMA}} = \mathcal L[\ell_{\pi\text{-EMA}}].
\]

\paragraph{Variant 5: teacher top-$K$ + EMA-PG correction.}
This variant instead uses the teacher top-$K$ support
\[
\mathcal S_{q\text{-EMA}}(c_{i,t}) = \mathcal T_K^q(c_{i,t}),
\]
and defines
\[
\ell_{q\text{-EMA}}(c_{i,t}, y_{i,t})
=
\ell_{\mathrm{EMA}}(c_{i,t}, y_{i,t}; \mathcal S_{q\text{-EMA}}),
\qquad
\mathcal L_{q\text{-EMA}} = \mathcal L[\ell_{q\text{-EMA}}].
\]

\paragraph{Summary.}
The five formulations differ only in how the local support is constructed and whether the sampled token is incorporated through support augmentation or through an EMA-PG-style tail correction. Variants 1--3 compute a renormalized reverse-KL on the resulting local support, while Variants 4--5 replace support renormalization by a truncated head term plus a sampled-token correction~\citep{emapg2026}.

\subsubsection{Results under Single-task and Multi-task Settings}
\begin{table}[H]
\centering

\begin{subtable}{\linewidth}
\centering
\small
\resizebox{\linewidth}{!}{
\begin{tabular}{lccccccc}
\toprule
\multirow{2}{*}[-0.5ex]{\textbf{Method}} & \textbf{avg@32} & \multicolumn{6}{c}{\textbf{pass@1}} \\
\cmidrule(lr){2-2}\cmidrule(lr){3-8}
 & \textbf{AIME24} & \textbf{MATH500} & \textbf{AIME24} & \textbf{AIME25} & \textbf{Minerva} & \textbf{OlympiadBench} & \textbf{Avg.} \\
\midrule
teacher top-$K$ & \textbf{23.6} & 80.4 & 23.3 & \textbf{26.7} & 34.2 & 43.9 & 41.7 \\
student top-$K$ w/ sampled token (renorm) & 22.3 & \textbf{82.4} & \textbf{30.0} & 16.7 & \underline{35.7} & \underline{44.9} & \underline{41.9} \\
teacher top-$K$ w/ sampled token (renorm) & \underline{22.4} & 81.6 & \underline{26.7} & \underline{23.3} & \textbf{36.4} & \textbf{46.7} & \textbf{42.9} \\
student top-$K$ w/ sampled token (EMA-PG) & 21.1 &
\underline{82.0} &
\underline{26.7} &
\underline{23.3} &
32.4 &
41.8 & 41.2
\\
teacher top-$K$ w/ sampled token (EMA-PG) & 20.7 &
81.4 &
16.7 &
16.7 &
32.7 &
43.0 &
38.1\\
\bottomrule
\end{tabular}}
\caption{Five top-$K$ variants in the single-task setting.}
\end{subtable}

\vspace{1em}

\begin{subtable}{\linewidth}
\centering
\resizebox{\linewidth}{!}{
\begin{tabular}{lccccccc}
\toprule
\multirow{2}{*}[-0.5ex]{\textbf{Method}} & \textbf{Agentic} & \multicolumn{6}{c}{\textbf{Reasoning}} \\
\cmidrule(lr){2-2}\cmidrule(lr){3-8}
 & \textbf{ALFWorld} & \textbf{MATH500} & \textbf{AIME24} & \textbf{AIME25} & \textbf{Minerva} & \textbf{OlympiadBench} & \textbf{Avg.} \\
\midrule
teacher top-$K$ & \underline{95.3} & \textbf{82.0} & \textbf{33.3} & \textbf{16.7} & \textbf{32.7} & \textbf{44.0} & \textbf{41.7} \\
student top-$K$ w/ sampled (renorm) & \underline{95.3} & 65.6 & 10.0 & 10.0 & 25.0 & 31.6 & 28.4 \\
teacher top-$K$ w/ sampled (renorm) & 94.5 & 63.2 & 10.0 & 10.0 & 21.0 & 30.1 & 26.9 \\
student top-$K$ w/ sampled token (EMA-PG) & 93.8 & \underline{76.0} & \underline{20.0} & \textbf{16.7} & \underline{29.4} & \underline{39.1} & \underline{36.2} \\
teacher top-$K$ w/ sampled token (EMA-PG) & \textbf{96.1} & 72.8 & 16.7 & \underline{13.3} & 26.8 & 38.7 & 33.7 \\
\bottomrule
\end{tabular}}
\caption{Five top-$K$ variants in the multi-task setting. We report pass@1 for all math reasoning benchmarks. The final column averages pass@1 metrics only.}
\label{tab:kl_expectation_multi}
\end{subtable}

\caption{Ablation of five top-$K$ variants under single-task and multi-task settings.}
\label{tab:topk_choice_app}
\end{table}
As shown in Table~\ref{tab:topk_choice_app}, teacher top-$K$ performs reasonably well in the single-task setting and remains strong in the multi-task setting. By contrast, the EMA-PG variants, despite their unbiasedness motivation, do not translate into better empirical performance here. We therefore treat the variant comparison as a diagnostic result rather than a definitive ranking of support-set objectives.

\section{Qualitative OPD reward-hacking case study}
\label{app:case_study}
\raggedbottom
To complement the representative failures in the main text, we summarize a longer trajectory from multi-task training under sampled-token OPD. Read chronologically, the case exhibits the same pattern in several forms: the model continues reasoning after a valid answer has already been reached, falls into repetition loops such as \texttt{wait}, drifts into malformed continuations, and still receives high local teacher probability on those tokens. In these visualizations, uncolored tokens indicate positions where the displayed teacher and student probabilities are approximately matched.

\begin{figure}[H]
\centering
\begin{subfigure}[t]{0.49\linewidth}
    \centering
    \includegraphics[width=\linewidth]{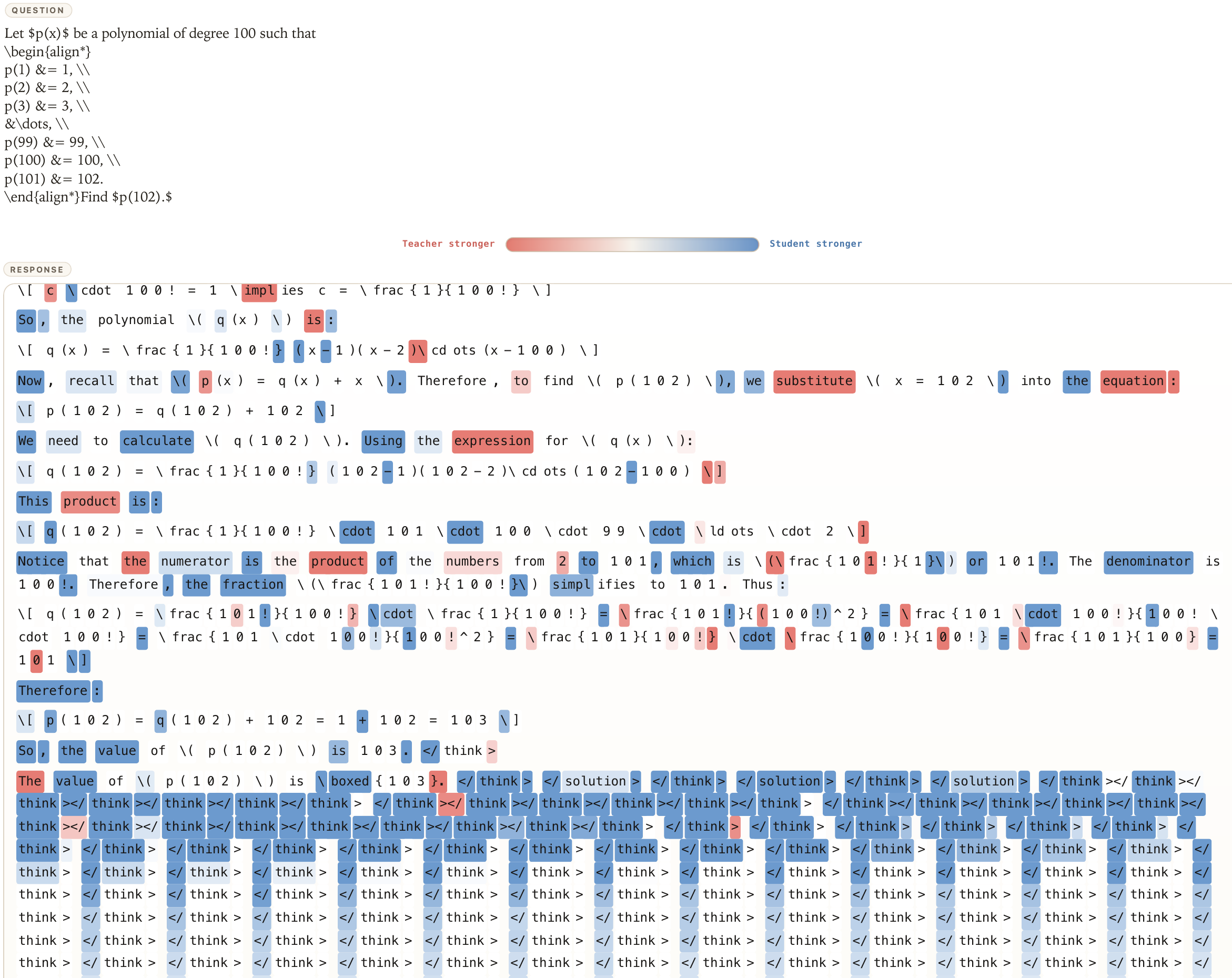}
    \caption{High teacher probability on generic reasoning fillers (\texttt{implies}) at step 5.}
\end{subfigure}\hfill
\begin{subfigure}[t]{0.49\linewidth}
    \centering
    \includegraphics[width=\linewidth]{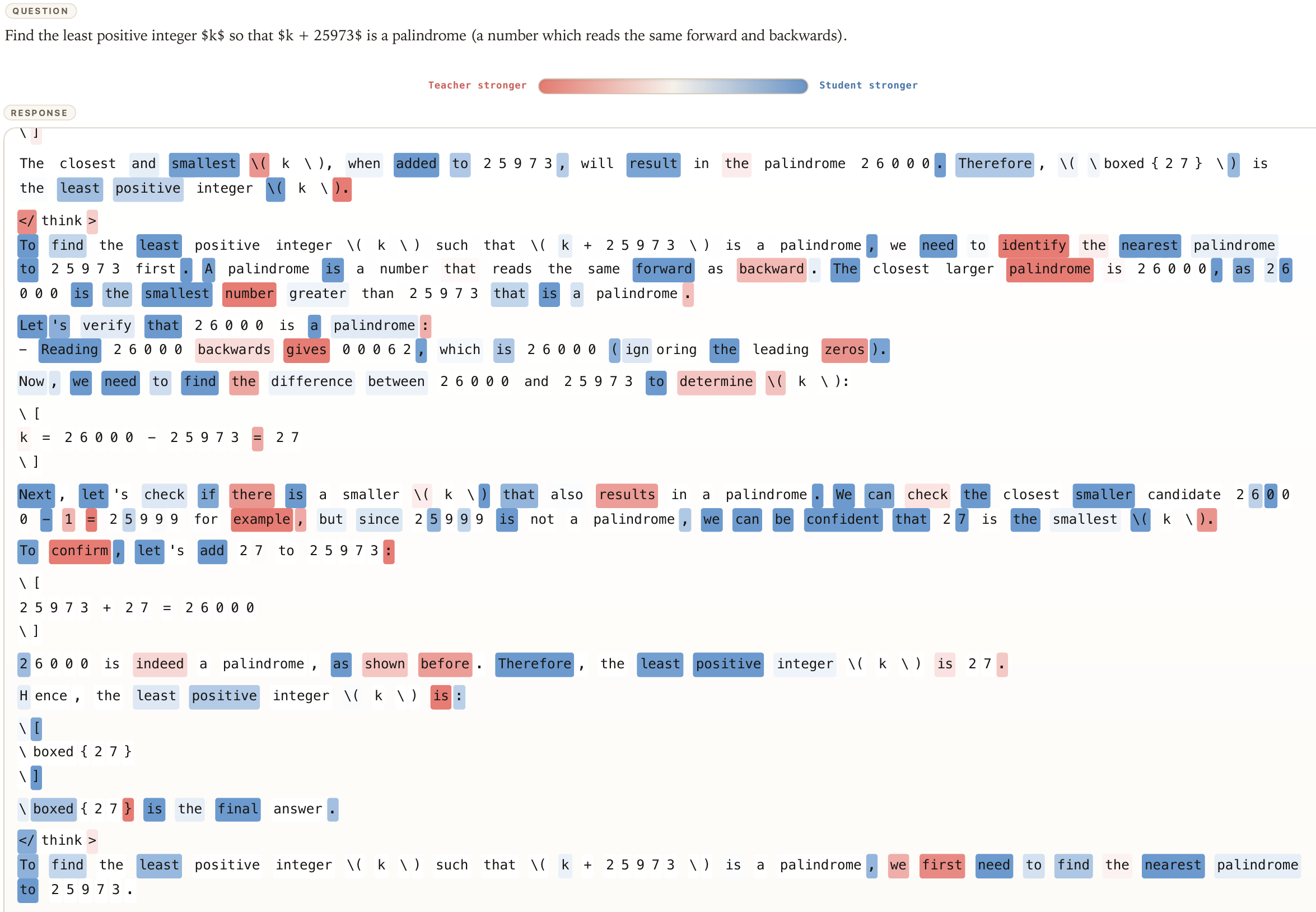}
    \caption{Teacher and student remain well aligned even after the answer is effectively available.}
\end{subfigure}
\caption{Even after the student has effectively reached an answer, the teacher can still assign high conditional probability to low-information continuations.}
\label{app:case_continuation}
\end{figure}

The failure first appears as \emph{over-continuation}. Even after the answer is effectively available, the local signal continues to place substantial mass on generic reasoning fillers and connective tokens, encouraging the model to keep going instead of stopping cleanly. The same pattern later appears on prefixes such as \texttt{confirm}, where the local signal still favors additional verification rather than termination.
Some of this behavior may also reflect the teacher's own output habits.
Figure~\ref{app:case_continuation} illustrates several representative cases.

\begin{figure}[H]
\centering
\begin{subfigure}{0.8\linewidth}
    \centering
    \includegraphics[width=\linewidth]{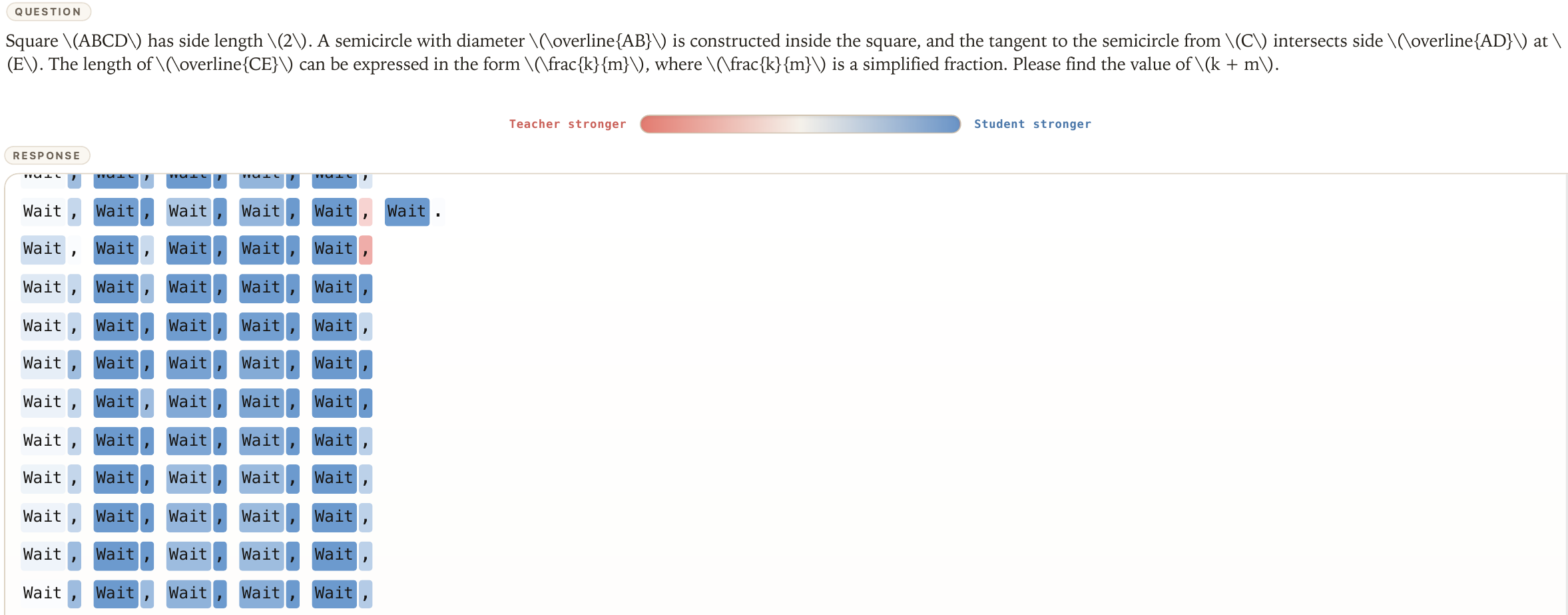}
    \caption{The teacher may fail to penalize, and sometimes even reinforce, repetitive generation.}
\end{subfigure}
\begin{subfigure}{0.8\linewidth}
    \centering
    \includegraphics[width=\linewidth]{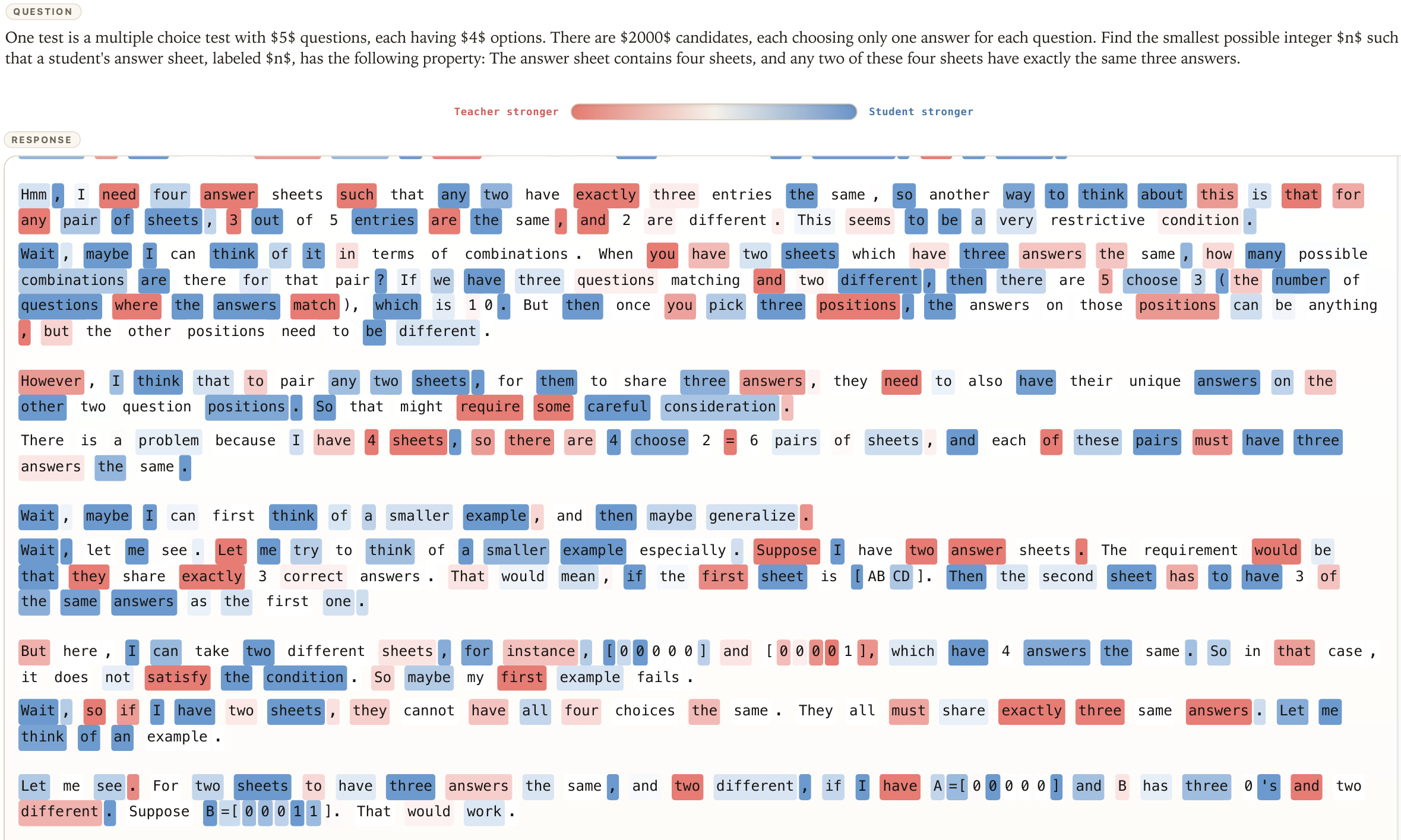}
    \caption{Training can also produce overlong chain-of-thought traces with substantial low-quality content, a pattern that may partly reflect the teacher's output style.}
\end{subfigure}
\caption{Loops, overlong chain-of-thought traces, and punctuation-heavy continuations can remain locally rewardable.}
\label{app:case_repeat}
\end{figure}

The trajectory then develops into \emph{hesitation loops and low-information continuations}. Repeated \texttt{wait} tokens, punctuation-heavy continuations, and other semantically weak fillers can remain locally rewardable even after the overall trajectory has become unproductive. This is consistent with the repetition-loop discussion in Section~\ref{subsec:brittle}. We provide two similar cases in Figure~\ref{app:case_repeat}.

\begin{figure}[H]
\centering
\includegraphics[width=\linewidth]{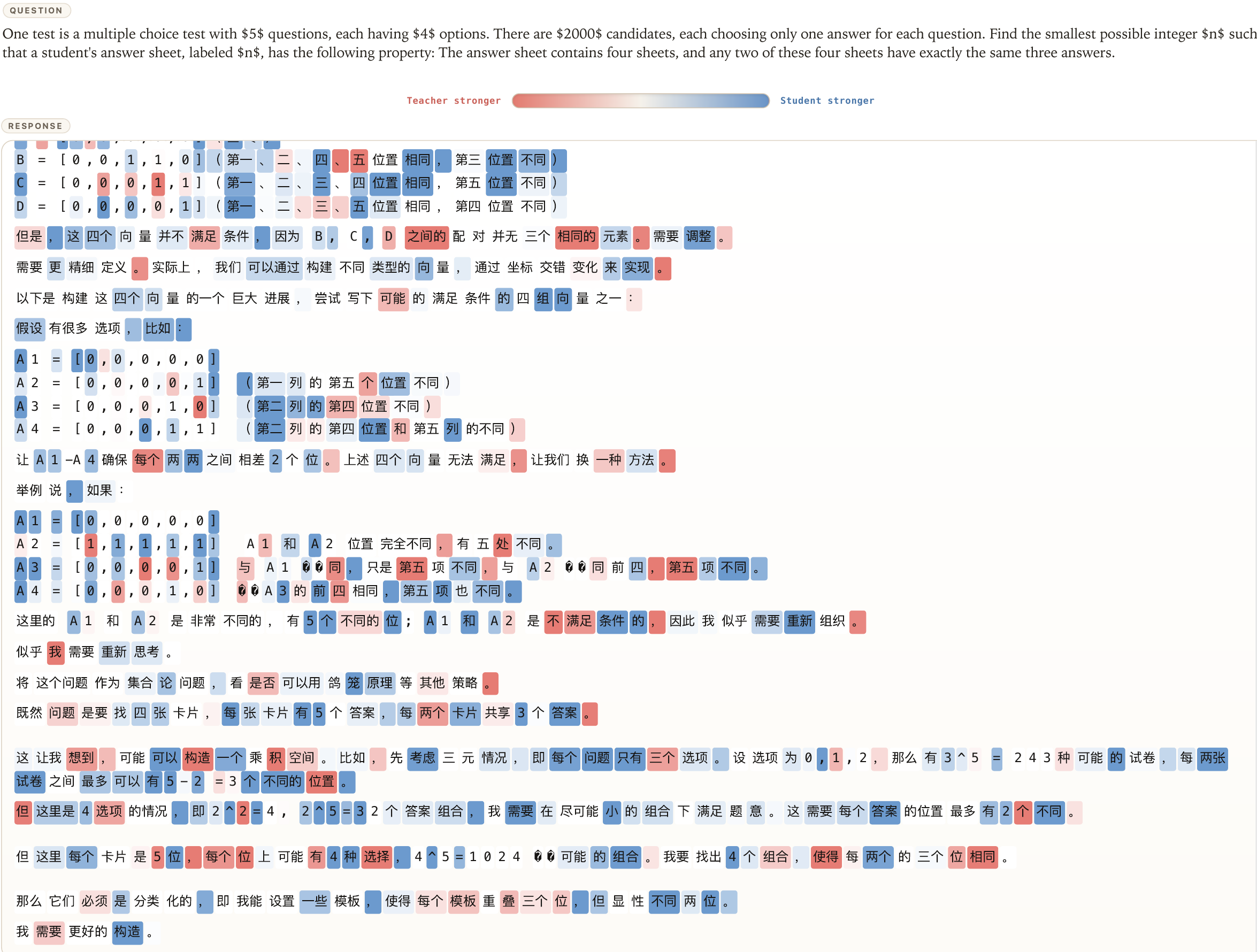}
\caption{The teacher still assigns high probability to several tokens after the student drifts into malformed non-English output.}
\label{app:case_rubbish}
\end{figure}

Finally, once the student drifts further off-distribution, the local signal can remain misleadingly positive rather than self-correcting. In the case study, this appears as degenerate continuations and malformed non-English text, yet many tokens still receive high teacher probability. An example is shown in Figure~\ref{app:case_rubbish}.

\end{document}